\DocumentMetadata{%
  lang        = en-US,
  pdfversion  = 1.7,
}

\def\arxivbuild{}
\ifdefined\arxivbuild
  \documentclass[11pt]{article}
  \usepackage{physbench-preprint}
\else
  \documentclass[lineno,nocopyright]{asmejour}
\fi

\ifdefined\arxivbuild
  \newcommand{\colfigw}{0.6\textwidth}    
  \newcommand{\coltabw}{0.67\textwidth}   
  \newcommand{\fwboxwidth}{100mm}         
\else
  \newcommand{\colfigw}{\columnwidth}
  \newcommand{\coltabw}{\columnwidth}
  \newcommand{\fwboxwidth}{62mm}
\fi

\allowdisplaybreaks

\usepackage{booktabs}   
\usepackage{graphicx}   
\usepackage{float}      
\usepackage{tikz}\usetikzlibrary{arrows.meta,positioning}  
\usepackage{algorithm}      
\usepackage{algpseudocode}  
\graphicspath{{figures/}}  

\usepackage{xcolor}

\newcommand{\benchname}{PhysicsBench}      
\newcommand{\sitebrand}{Engineering AI Leaderboard} 
\newcommand{\siteurl}{leaderboard.narnia.ai}

\hypersetup{allcolors=black, linkcolor=black, citecolor=black, urlcolor=black,
            filecolor=black, anchorcolor=black, menucolor=black, runcolor=black}
\ifdefined\linenumberfont
  \renewcommand{\linenumberfont}{\normalfont\scriptsize}
\fi

\ifdefined\arxivbuild
  \definecolor{arxivlink}{RGB}{0,60,150}
  \hypersetup{colorlinks=true, allcolors=arxivlink, linkcolor=arxivlink,
              citecolor=arxivlink, urlcolor=arxivlink, filecolor=arxivlink,
              anchorcolor=arxivlink, menucolor=arxivlink, runcolor=arxivlink}
\fi

\hypersetup{%
  pdfauthor   = {Sang Won Lee, Hyogu Jeong, Namwoo Kang},
  pdftitle    = {PhysicsBench: A Unified Leaderboard for Generative and Predictive Models in Engineering Design and Simulation},
  pdfkeywords = {Benchmark, Engineering design, Computer-aided engineering, Geometry generation, Surrogate modeling, Neural operators, Small-data regime},
  pdfsubject  = {A benchmark for generative and predictive models in engineering design and simulation},
}

\JourName{Mechanical Design}

\begin{document}

\ifdefined\arxivbuild
  \physbAffiliation{a}{Narnia Labs, Daejeon, Republic of Korea}
  \physbAffiliation{b}{Cho Chun Shik Graduate School of Mobility, KAIST,
    Daejeon, Republic of Korea}
  \physbAuthorAffiliation{1}{a}{}
  \physbAuthorAffiliation{2}{a}{}
  \physbAuthorAffiliation{3}{a,b}{nwkang@kaist.ac.kr}
\fi

\SetAuthorBlock{Sang Won Lee\textsuperscript{\dag}}{%
  Narnia Labs,\\
  Daejeon, Republic of Korea}

\SetAuthorBlock{Hyogu Jeong\textsuperscript{\dag}}{%
  Narnia Labs,\\
  Daejeon, Republic of Korea}

\SetAuthorBlock{Namwoo Kang\CorrespondingAuthor}{%
  Narnia Labs, Daejeon, Republic of Korea\\
  Cho Chun Shik Graduate School of Mobility, KAIST,\\
  Daejeon, Republic of Korea\\
  email: nwkang@kaist.ac.kr}

\title{\benchname{}: A Unified Leaderboard for Generative and Predictive Models in Engineering Design and Simulation}

\keywords{Benchmark, Engineering design, Computer-aided engineering, Geometry generation, Surrogate modeling, Neural operators, Small-data regime}

\begin{abstract}
Generative and predictive artificial intelligence models are increasingly used
to generate geometry and to predict physical fields and scalar quantities in
engineering design and simulation. Yet these models are typically evaluated in isolation, on
academic datasets at unconstrained scales, with inconsistent metrics and
procedures. We present \benchname{}, a unified benchmark
and leaderboard that evaluates generative and predictive models under
one standardized procedure.
\benchname{} spans seven generation and prediction tasks across 1D, 2D, and 3D
domains and ranks 66 models on nine datasets, comprising industrial-scale
CAD/CFD/FEA simulations and public references, expanded into 28 configurations.
One procedure and ranking apply
to both families, each ranked within its own tasks. Evaluation spans
realistic, limited data scales from S to XL rather than the unlimited
training sets common in academic benchmarks. A common metric suite captures
geometric fidelity with distributional distances, physical-field and scalar
accuracy, and engineering-specific field- and shape-validity. BenchRank
debiases correlated metrics and ranks by PageRank
over a head-to-head dominance graph, so every reported quality metric is also
ranked, with computational cost in a separate efficiency view.
Across tasks, an architecture's large-scale academic standing weakly predicts
its small-data ranking. The top model changes with data scale in six of the
seven tasks, and no model leads more than one task. \benchname{}
turns ``state-of-the-art'' from a self-reported claim into an openly published
foundation for model selection.
\end{abstract}

\date{}

\maketitle

\revfootnote{\textsuperscript{\dag}These authors contributed equally to this work.}

\section{Introduction}
\label{sec:intro}

Numerical simulation has long served as a cornerstone of modern engineering design, enabling the prediction of physical behavior across disciplines ranging from structural mechanics to fluid dynamics. High-fidelity methods such as Computational Fluid Dynamics (CFD) and Finite Element Analysis (FEA) provide accurate performance estimates. Their reliance on fine spatial and temporal discretizations, however, renders them computationally expensive, so a single design evaluation frequently demands specialized software, domain expertise, and extensive high-performance computing resources~\cite{forrester2009recent}. This cost becomes prohibitive within iterative design and optimization loops, where thousands of candidate geometries must be assessed. To alleviate this burden, data-driven surrogate models based on deep learning have recently emerged as a compelling alternative, learning the mapping from design geometry to physical response directly from simulation data and thereby delivering near-instantaneous predictions~\cite{lu2021deeponet,li2021fno}. In particular, advances in geometric deep learning, neural operators, and transformer-based neural solvers have accelerated aerodynamic and structural analyses by several orders of magnitude while retaining engineering-level accuracy~\cite{qi2017pointnet,kovachki2023neuraloperator,wu2024transolver}. Beyond such forward prediction, generative models have opened a complementary, inverse approach to design, synthesizing novel and manufacturable geometries that satisfy prescribed performance targets. Generative Adversarial Networks (GANs), Variational Autoencoders (VAEs), normalizing flows, and diffusion models extend Artificial Intelligence (AI) from the evaluation of designs to their creation~\cite{goodfellow2014gan,kingma2014vae,yang2019pointflow,ho2020ddpm}. Combined, these generative and predictive capabilities are reshaping the engineering design process, expanding the design space that can be explored and enabling real-time, interactive evaluation of complex three-dimensional systems~\cite{regenwetter2022review,kang2025scenarios}.

The rapid progress of AI has been inseparable from the availability of large-scale, publicly ranked benchmarks. In computer vision and natural language processing, standardized datasets and leaderboards such as ImageNet~\cite{deng2009imagenet,russakovsky2015imagenet}, GLUE~\cite{wang2019glue}, SuperGLUE~\cite{wang2019superglue}, and MMLU~\cite{hendrycks2021mmlu} established common evaluation protocols that enabled fair comparison, ensured reproducibility, and catalyzed successive breakthroughs across competing model families~\cite{ott2022mapping}. Motivated by this precedent, the scientific machine learning community has begun to construct analogous benchmarks for physics-based simulation. PDEBench~\cite{pdebench} provides standardized datasets and baselines spanning a broad range of time-dependent Partial Differential Equations (PDEs), whereas the Well~\cite{ohana2024thewell} assembles a 15 TB collection of diverse physical simulations, ranging from fluid dynamics to acoustics and astrophysics, under a single evaluation interface. Within engineering design specifically, large-scale CFD datasets such as AirfRANS~\cite{airfrans} for airfoil aerodynamics and DrivAerNet++~\cite{elrefaie2024drivaernet} for three-dimensional automotive aerodynamics have introduced public leaderboards for surrogate-model evaluation. CarBench~\cite{elrefaie2025carbench} has recently established the first comprehensive and reproducible benchmark ranking eleven state-of-the-art architectures on high-fidelity car aerodynamics. Such ranked leaderboards align evaluation practices across the community, quantify genuine algorithmic progress, and lower the barrier to entry for new methods, thereby accelerating the maturation of data-driven engineering.

Despite this progress, several research gaps remain unaddressed. Most existing benchmarks are narrow in scope, targeting a single physical regime or a canonical reference geometry, and many still rely on simplified two-dimensional cases or idealized three-dimensional shapes that omit the geometric complexity of production components~\cite{pdebench,airfrans}. Their coverage is further constrained by a scarcity of high-fidelity, experimentally validated data and by non-standardized train, validation, and test splits, which obscure fair comparison and hinder the assessment of true generalization to unseen designs~\cite{kapoor2022data}. Moreover, prevailing benchmarks are calibrated to the large-data conditions of mainstream machine learning, implicitly assuming access to thousands of labelled simulations. High-fidelity CFD and FEA campaigns in industrial practice, by contrast, yield only a limited number of samples per design study, leaving the low-data regime that dominates engineering deployment largely unexamined. Consequently, current leaderboards capture algorithmic capability under idealized settings, yet reveal little about model behavior under the constraints that determine practical utility.

Motivated by these limitations, this work introduces \benchname{}, which adopts the public leaderboard model but corrects the shortcomings identified above. \benchname{} brings this methodology to engineering design, where no comparable leaderboard yet exists, evaluating the generation of candidate geometry and the prediction of its physical
performance, the two stages conventionally served by design authoring and numerical
solvers, under the conditions that govern engineering practice rather than academic
convenience. It is a unified benchmark and leaderboard, deployed as the \sitebrand{} at \url{https://\siteurl}, that assesses both generative and predictive models under a single standardized procedure of training, inference, metric
computation, and ranking on industrial engineering datasets. By applying this procedure uniformly, \benchname{} isolates genuine algorithmic differences from differences in evaluation setup, yielding rankings that reflect a common, independently applied standard rather than self-selected reporting conditions.

\benchname{} rests on two principles. The first is a single standardized procedure and ranking applied uniformly across all generative and predictive tasks. The second is evaluation across realistic and limited data scales, reported as small (S), medium (M), large (L), and extra large (XL) buckets, rather than the
unlimited training sets common in academic benchmarks. The benchmark spans seven tasks across 1D, 2D, and 3D domains, supported by a common metric suite covering geometric fidelity, including distributional distances, physical-field and scalar accuracy, engineering-specific field and shape-validity, and computational cost. Rankings are produced by BenchRank, a debiased graph-based procedure in which each quality metric contributes directly to the ranking, while computational cost is reported separately in an efficiency view. The goal of \benchname{} is not to identify a single best model but to provide decision support, identifying for each task and data scale the best-performing models on a given dataset.

\benchname{} makes five contributions.
\begin{itemize}
  \item \textbf{A unified cross-family leaderboard.} Prior engineering-design
        frameworks such as EngiBench, proposed by Felten et
        al. \cite{felten2025engibench}, provide common APIs, curated datasets,
        and modular support for generative, surrogate, and optimization methods,
        establishing a valuable foundation for reproducible comparison on
        engineering design problems. \benchname{} extends this direction toward
        model-output evaluation at industrially relevant scales by ranking
        both generative and predictive models under a single debiased leaderboard
        and across a controlled range of data scales. It spans seven generation
        and prediction tasks across 1D, 2D, and 3D engineering domains, on
        CAD/CFD/FEA data ranging from industrial-scale simulations to established
        public references. Each family is ranked within its own tasks and metric
        suite, and \benchname{} does not rank generators against predictive models
        head-to-head.
  \item \textbf{Realistic-scale evaluation.} Every model is swept across
        controlled data scales from S to XL. This sweep exposes data-efficiency
        curves, small-data collapse, and best-model crossovers that fixed-scale
        academic evaluation hides, and it shows that a model's large-scale
        academic standing is a weak predictor of its ranking in the small-data
        regime.
  \item \textbf{BenchRank.} \benchname{} introduces a debiased, graph-based
        ranking purpose-built for multi-metric engineering evaluation. BenchRank
        down-weights correlated metrics, ranks by PageRank over a head-to-head
        dominance graph, gates non-viable models, and aggregates across datasets
        by a geometric mean. The procedure is deterministic and reproducible.
  \item \textbf{Ranked engineering-validity metrics.} Beyond pixel and
        point-averaged error, \benchname{} ranks \emph{field-validity} for
        prediction and \emph{shape-validity} for generation. Field-validity
        covers directional and modal alignment measured by the modal assurance
        criterion, together with pattern fidelity measured by sign and extremal
        agreement. Shape-validity covers structural manifold quality measured by
        the manifold-$\Delta$ and uniformity-$\Delta$ scores. A model that is
        metrically close but physically invalid is therefore not rewarded.
  \item \textbf{A reproducible, living platform.} Every published ranking is
        deterministically regenerable from the committed per-run data,
        contributors add new models without conflicts, and the leaderboard
        exposes the underlying per-run scores.
\end{itemize}

The remainder of this paper is organized as follows. Section \ref{sec:related} reviews generative models, predictive models, and engineering benchmarks. Section \ref{sec:method} describes the \benchname{} datasets, tasks, pipeline, metrics, and BenchRank. Section \ref{sec:results} presents and discusses the leaderboard results, and Section \ref{sec:conclusion} concludes with limitations and future work.

\section{Related Work}
\label{sec:related}

\subsection{Generative Models for Engineering Design}

Deep generative models synthesize candidate engineering designs directly from data. In 2D, Generative Adversarial Networks (GANs), Variational Autoencoders (VAEs), and denoising diffusion models \cite{ho2020ddpm} generate engineering images and physical fields discretized on regular grids. In 3D, implicit coordinate networks (DeepSDF \cite{park2019deepsdf}), continuous normalizing flows (PointFlow \cite{yang2019pointflow}), and point-cloud diffusion models produce shapes as continuous implicit surfaces or discrete point clouds. Regenwetter et al. \cite{regenwetter2022review} review these methods for engineering design. Several domain-specific generative-design datasets have been released, including BIKED for bicycles \cite{regenwetter2022biked}, DeepCAD for CAD construction sequences \cite{wu2021deepcad}, and Ship-D for ship hulls \cite{bagazinski2023shipd}. These datasets target shape or CAD synthesis
within a single domain and include no surrogate prediction task.

Generative models are benchmarked on academic shape collections such as ShapeNet~\cite{chang2015shapenet} using distributional metrics such as the Fréchet Inception Distance (FID), the Chamfer distance, and the coverage (COV) and minimum-matching-distance (MMD) suite of Achlioptas et al.~\cite{achlioptas2018pointcloud}. These metrics measure sample fidelity and diversity. They do not test whether generated geometry lies on a valid, evenly sampled shape manifold. \benchname{} promotes this structural test to a ranked pair of shape-validity metrics, manifold-$\Delta$ and uniformity-$\Delta$. Two gaps remain. Generative models are rarely evaluated on industrial CAD geometry under real data-scarcity and manufacturability constraints, and their behavior at engineering-scale data size is uncharacterized.

\subsection{Neural Surrogate and Operator-Learning Models}

Neural surrogates replace expensive solvers with learned models that predict physical fields or scalar quantities. Neural operators learn mappings between
function spaces and generalize across discretizations, including the Fourier Neural Operator \cite{li2021fno}, DeepONet \cite{lu2021deeponet}, and the broader
operator-learning frameworks \cite{kovachki2023neuraloperator}. Geometry-aware operators and transformers extend this to irregular engineering geometry,
including the Geometry-Informed Neural Operator~\cite{li2023gino}, Transolver \cite{wu2024transolver}, GNOT \cite{hao2023gnot}, and universal physics transformers \cite{alkin2024upt} together with their anchored-branched variant for automotive CFD (AB-UPT) \cite{alkin2025abupt}. Graph-network simulators \cite{sanchezgonzalez2020gns,pfaff2021meshgraphnets} and point and graph-based backbones such as PointNet \cite{qi2017pointnet} and its successors
operate directly on meshes and point clouds. A complementary line imposes physical laws during training, including physics-informed neural networks \cite{raissi2019pinn} and the physics-informed neural operator \cite{li2021pino}, which learn from data while respecting the governing equations. Image-based surrogates regress fields on rasterized geometry using encoder--decoder CNNs such as U-Net \cite{ronneberger2015unet}, feature-pyramid networks \cite{lin2017fpn}, and dense-prediction vision transformers \cite{ranftl2021dpt,xie2021segformer}, and map images to scalar quantities using CNN and vision-transformer backbones \cite{he2016resnet,touvron2021deit}. For low-dimensional tabular and time-series quantities of interest, surrogates draw on gradient-boosted trees, deep tabular networks, and in-context foundation models \cite{gorishniy2021ftt,arik2021tabnet,hollmann2025tabpfn}, together with recurrent and convolutional sequence models \cite{hochreiter1997lstm,bai2018tcn}.

Each model is typically reported on its own datasets with method-specific metrics and training budgets. This precludes cross-method and cross-dataset comparison and leaves the generative and predictive literatures disconnected, despite both serving the same design loop.

\subsection{Benchmarks and Leaderboards in Machine Learning}
\label{sec:related-leaderboards}

Standardized benchmarks with public rankings have driven progress in machine learning, from ImageNet \cite{deng2009imagenet,russakovsky2015imagenet} and
GLUE/SuperGLUE \cite{wang2019glue,wang2019superglue} to MLPerf \cite{mattson2020mlperf,reddi2020mlperf} and HELM \cite{liang2022holistic}, with
community platforms \cite{paperswithcode,kaggle,open_llm_leaderboard} extending the format to thousands of tasks.

Such rankings, however, exhibit well-characterized failure modes. They depend on task selection and aggregation \cite{dehghani2021benchmark}, and a high score on a narrow task does not certify general capability \cite{raji2021everything}. Static benchmarks also saturate as models catch up to their hardest tasks. On GLUE, models passed the non-expert human baseline within a year, prompting SuperGLUE \cite{wang2019superglue}, and a survey of 3{,}765 benchmarks reports rapid saturation \cite{ott2022mapping}, motivating dynamic evaluation \cite{kiela2021dynabench}. Test-set contamination inflates scores through memorization rather than generalization \cite{sainz-etal-2023-nlp}. Optimizing one proxy metric is a known failure mode \cite{thomas2022reliance}. Rank diverges from deployment utility when cost is omitted \cite{ethayarajh-jurafsky-2020-utility}, and efficiency is mis-measured when reduced to one indicator \cite{dehghani2022efficiency}. Results are often irreproducible without committed code, data, and configuration \cite{pineau2021improving}. \benchname{} answers each failure mode directly. BenchRank down-weights correlated metrics and ranks on a metric suite rather than one number. Computational cost is reported in a separate efficiency view. The leaderboard is regenerable from committed per-model results on fixed held-out splits. Rankings are reported per task and per data scale, not as one universal score.

The small-data regime is where engineering operates and where rankings are least stable. High-fidelity CFD/FEA evaluation is the dominant cost of surrogate modeling, therefore the industrial surrogates train on tens to a few hundred simulations \cite{forrester2009recent,kapoor2022data}, and recent design methods operate directly in this data-constrained regime \cite{kwon2025shapeopt}. The best model changes with data scale. On tabular data, gradient-boosted trees lead at larger sizes, while an in-context foundation model wins on the smallest training sets \cite{grinsztajn2022tree,mcelfresh2023when,hollmann2025tabpfn}. A leaderboard built only at one specific scale is a poor guide to the regime engineers face. \benchname{} therefore sweeps controlled S/M/L/XL data scales rather than reporting at a single fixed size.

\subsection{Engineering and Scientific-ML Benchmarks}

Several engineering benchmarks standardize complementary aspects of evaluation. PDEBench \cite{pdebench} and The Well \cite{ohana2024thewell} assemble large collections of time-dependent PDE simulations for operator-learning benchmarks. AirfRANS \cite{airfrans} provides RANS airfoil flow, while DrivAerNet++ \cite{elrefaie2024drivaernet} and DrivAerML \cite{ashton2024drivaerml} provide large-scale automotive-aerodynamics datasets with deep-learning surrogate benchmarks. Concurrent with our work, Elrefaie et al. \cite{elrefaie2025carbench} introduced CarBench, a strong domain-specific benchmark for 3D automotive aerodynamics that enables reproducible comparison of eleven neural surrogates on DrivAerNet++ across accuracy, physical consistency, efficiency, bootstrap uncertainty, and Pareto trade-offs. This depth of evaluation provides an important reference for model assessment within a high-fidelity engineering domain. EngiBench \cite{felten2025engibench} complements this approach by establishing valuable open-source infrastructure for reproducible data-driven engineering design research through a common API, curated cross-domain problems and datasets, and modular support for generative, surrogate, and optimization methods. Building on these complementary foundations, a remaining need is to compare generative and predictive models across a broader range of engineering settings, including 1D, 2D, and 3D generation and prediction, under the limited training data budgets typical of industrial simulation campaigns. \benchname{} addresses this need through controlled data-scale evaluation on diverse CAD/CFD/FEA tasks using a common multi-metric ranking procedure.

\subsection{Industrial CAE Datasets}
\benchname{} builds on publicly released, simulation-backed engineering datasets, including the jet-engine-bracket structural dataset DeepJEB \cite{hong2025deepjeb} and the synthetic wheel dataset DeepWheel \cite{yoo2025deepwheel}, which extends aforementioned generative wheel design and evaluation \cite{yoo2021wheel}, the automotive-aerodynamics datasets DrivAerNet++ \cite{elrefaie2024drivaernet} and DrivAerML \cite{ashton2024drivaerml}, the PDE-simulation benchmark PDEBench \cite{pdebench}, and standard tabular and time-series references. These datasets supply the real geometries, coupled multi-physics fields, and engineering quantities of interest that separate industrial evaluation from simplified academic data.

\section{Benchmark}
\label{sec:method}

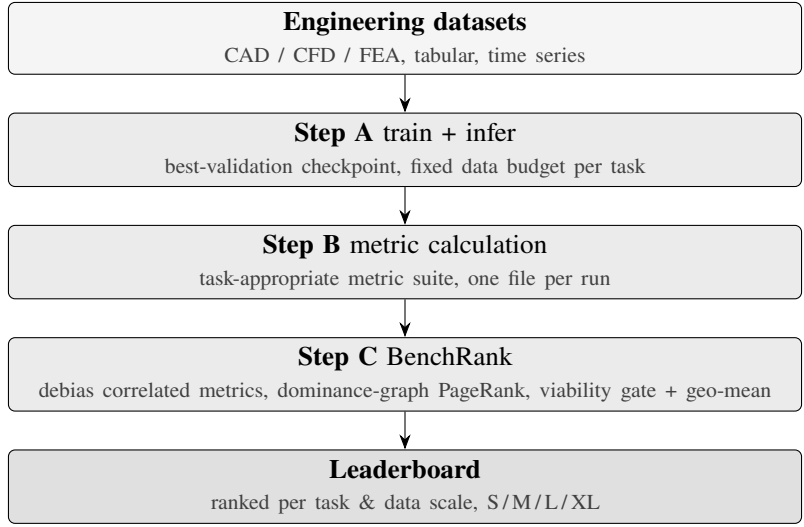
\begin{figure}[t]
  \centering
  \newcommand{\fgsub}[1]{{\scriptsize\textcolor{black!75}{#1}}}%
  \begin{tikzpicture}[
      font=\small,
      >={Stealth[length=2mm]},
      pbox/.style={draw, rounded corners=2.5pt, align=center,
                   inner xsep=7pt, inner ysep=3.5pt, minimum height=6mm, text width=\fwboxwidth},
      dsbox/.style={pbox, fill=black!4},
      stbox/.style={pbox, fill=black!8},
      obox/.style={pbox, fill=black!12}]
    \node[dsbox] (ds)
      {\textbf{Engineering datasets} \\ \fgsub{CAD / CFD / FEA, tabular, time series}};
    \node[stbox, below=5mm of ds] (a)
      {\textbf{Step A} train + infer \\ \fgsub{best-validation checkpoint, fixed data budget per task}};
    \node[stbox, below=5mm of a] (b)
      {\textbf{Step B} metric calculation \\ \fgsub{task-appropriate metric suite, one file per run}};
    \node[stbox, below=5mm of b] (c)
      {\textbf{Step C} BenchRank \\ \fgsub{debias correlated metrics, dominance-graph PageRank, viability gate + geo-mean}};
    \node[obox, below=5mm of c] (lb)
      {\textbf{Leaderboard} \\ \fgsub{ranked per task \& data scale, S\,/\,M\,/\,L\,/\,XL}};
    \draw[->] (ds) -- (a);  \draw[->] (a) -- (b);
    \draw[->] (b) -- (c);   \draw[->] (c) -- (lb);
  \end{tikzpicture}
  \caption[Overview of the \benchname{} evaluation pipeline]%
  {Overview of the \benchname{} evaluation pipeline. The datasets combine
  industrial CAD/CFD/FEA simulation sets with established public references.}
  \label{fig:framework}
\end{figure}

\benchname{} evaluates generative and predictive models under a single standardized procedure that fixes the data, training, inference, metric computation, and ranking applied to every model. This section formalizes that procedure and the debiased leaderboard it produces. Figure~\ref{fig:framework} presents an overview of the \benchname{} evaluation pipeline.

\subsection{Datasets and Task Taxonomy}
\label{sec:method:data}

\begin{figure*}[t]
  \centering
  \includegraphics[width=\textwidth]{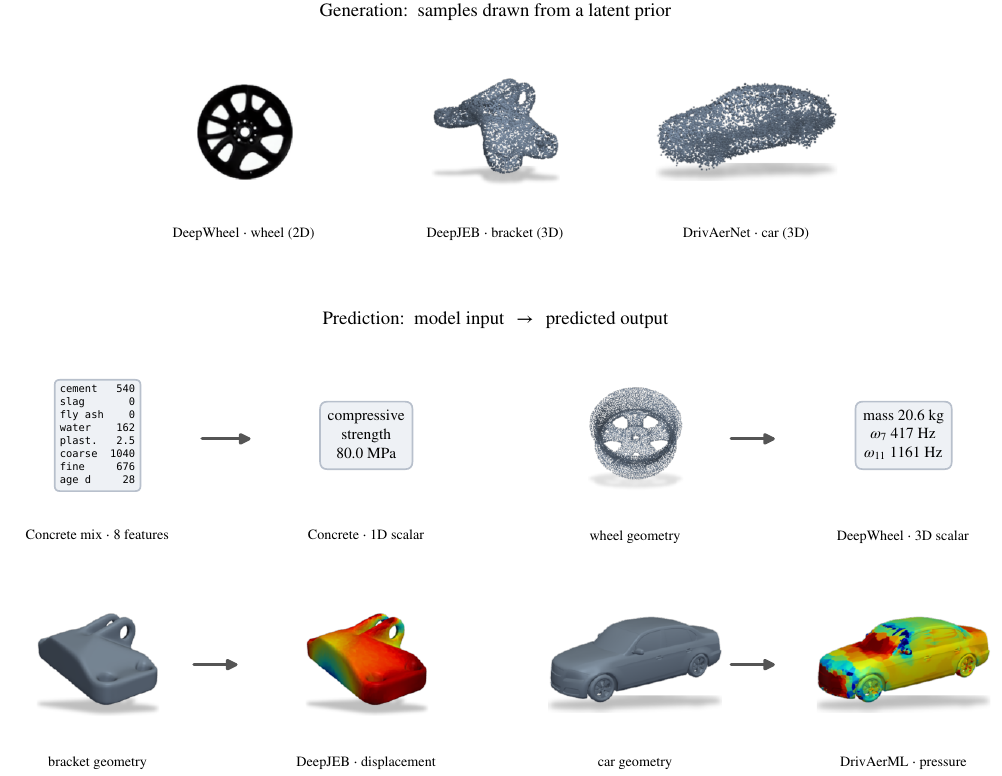}
  \caption[Representative tasks across \benchname{}]%
  {Representative tasks across \benchname{}, ordered as in
  Table~\ref{tab:datasets} by dimension and, at the top level, generation before
  prediction. Geometry is shaded neutral grey, whether generated or given as model
  input, and predicted fields use the jet colormap. The 2D field-prediction
  tasks are shown separately with the qualitative results in
  Section~\ref{sec:results:pertask}.}
  \label{fig:gallery}
\end{figure*}

\benchname{} draws on engineering CAD/CFD/FEA datasets, both industrial-scale simulation sets and established public references, spanning structural, modal, aerodynamic, porous-flow, acoustic, prognostic, and materials disciplines. Every dataset name follows a four-token convention and decomposes into five
orthogonal facets. The first is the input modality dimension. The second is the \emph{output-support manifold} dimension, where a scalar maps to \texttt{1d}, an image, plane, or surface 2-manifold maps to \texttt{2d}, and a volume 3-manifold maps to \texttt{3d}. The third is the engineering scenario,
namely the load case, region, or phenomenon that distinguishes siblings. The fourth is the discipline. The fifth is \emph{coupling}, meaning whether one solver run produces the output jointly. We adopt the principle that one engineering problem maps to exactly one dataset and one model-scale checkpoint. A model emits the problem's coupled multi-dimensional output in a single forward pass, and the leaf quantities, for example pressure and velocity for an aerodynamic field or displacement and stress for a structural load case, are stored as \emph{components} of that one result rather than as separate training runs. Components
and their vector groupings are defined consistently across the benchmark. Table~\ref{tab:datasets} lists
the datasets grouped by the seven tasks, and Figure~\ref{fig:gallery} illustrates representative tasks as input-to-output pairs, spanning generation and prediction across 1D, 2D, and 3D on real engineering data.

\begin{table*}[t]
  \centering
  \footnotesize
  \caption[Datasets in \benchname{}, grouped by task]%
  {Datasets in \benchname{}, grouped by task. ``Identifier'' is the
  four-token name each dataset is published and released under. ``In$\to$Out''
  gives the input-modality and output-support manifold dimensions, and ``QoI''
  lists the coupled quantities of interest as components, with vector quantities
  contributing one component per axis. Generation tasks have no target field, so
  their QoI entry is the geometry itself.}
  \label{tab:datasets}
  \begin{tabular}{@{}l l l c l l@{}}
\toprule
Task & Dataset & Identifier & In$\to$Out & QoI (components) & License \\
\midrule
1D Scalar Prediction & UCI-Airfoil Self-Noise & \texttt{airfoil\_1d\_1d\_noise} & 1d$\to$1d & sound pressure & CC BY 4.0 \\
 & NASA-CMAPSS RUL & \texttt{cmapss\_1dt\_1d\_rul} & 1dt$\to$1d & remaining useful life & Public Domain \\
 & UCI-Concrete Strength & \texttt{concrete\_1d\_1d\_strength} & 1d$\to$1d & compressive strength & CC BY 4.0 \\
\addlinespace
2D Image Generation & DeepJEB & \texttt{deepjeb\_2d\_2d} & 2d$\to$2d & geometry & ODC-By v1.0 \\
 & DeepWheel & \texttt{deepwheel\_2d\_2d} & 2d$\to$2d & geometry & CC BY-NC 4.0 \\
 & DrivAerNet & \texttt{drivaernet\_2d\_2d} & 2d$\to$2d & geometry & CC BY-NC 4.0 \\
\addlinespace
2D Scalar Prediction & DeepJEB Structural & \texttt{deepjeb\_2d\_1d\_structural} & 2d$\to$1d & displacement, stress & ODC-By v1.0 \\
 & DeepWheel Mass & \texttt{deepwheel\_2d\_1d\_mass} & 2d$\to$1d & mass & CC BY-NC 4.0 \\
 & DeepWheel Modal & \texttt{deepwheel\_2d\_1d\_modal} & 2d$\to$1d & natural frequencies & CC BY-NC 4.0 \\
\addlinespace
2D Field Prediction & AirfRANS Flow & \texttt{airfrans\_2d\_2d\_flow} & 2d$\to$2d & velocity, pressure, turbulence & ODbL 1.0 \\
 & DeepJEB Structural & \texttt{deepjeb\_2d\_2d\_structural} & 2d$\to$2d & displacement, stress & ODC-By v1.0 \\
 & DeepWheel Depth & \texttt{deepwheel\_2d\_2d\_depth} & 2d$\to$2d & depth & CC BY-NC 4.0 \\
 & PDEBench Darcy Pressure & \texttt{pdebenchdarcy\_2d\_2d\_pressure} & 2d$\to$2d & pressure & CC BY 4.0 \\
\addlinespace
3D Geometry Generation & DeepJEB & \texttt{deepjeb\_3d\_3d} & 3d$\to$3d & geometry & ODC-By v1.0 \\
 & DeepWheel & \texttt{deepwheel\_3d\_3d} & 3d$\to$3d & geometry & CC BY-NC 4.0 \\
 & DrivAerNet & \texttt{drivaernet\_3d\_3d} & 3d$\to$3d & geometry & CC BY-NC 4.0 \\
\addlinespace
3D Scalar Prediction & DeepJEB Structural & \texttt{deepjeb\_3d\_1d\_structural} & 3d$\to$1d & displacement, stress & ODC-By v1.0 \\
 & DeepWheel Mass & \texttt{deepwheel\_3d\_1d\_mass} & 3d$\to$1d & mass & CC BY-NC 4.0 \\
 & DeepWheel Modal & \texttt{deepwheel\_3d\_1d\_modal} & 3d$\to$1d & natural frequencies & CC BY-NC 4.0 \\
\addlinespace
3D Field Prediction & DeepJEB Diagonal Load & \texttt{deepjeb\_3d\_2d\_diagonal} & 3d$\to$2d & displacement, stress & ODC-By v1.0 \\
 & DeepJEB Horizontal Load & \texttt{deepjeb\_3d\_2d\_horizontal} & 3d$\to$2d & displacement, stress & ODC-By v1.0 \\
 & DeepJEB Modal & \texttt{deepjeb\_3d\_2d\_modal} & 3d$\to$2d & mode shapes & ODC-By v1.0 \\
 & DeepJEB Torsional Load & \texttt{deepjeb\_3d\_2d\_torsion} & 3d$\to$2d & displacement, stress & ODC-By v1.0 \\
 & DeepJEB Vertical Load & \texttt{deepjeb\_3d\_2d\_vertical} & 3d$\to$2d & displacement, stress & ODC-By v1.0 \\
 & DrivAerML Surface & \texttt{drivaerml\_3d\_2d\_surface} & 3d$\to$2d & pressure, wall shear stress & CC BY-SA 4.0 \\
 & DrivAerML Volume & \texttt{drivaerml\_3d\_3d\_volume} & 3d$\to$3d & p-coeff, velocity, vorticity & CC BY-SA 4.0 \\
 & DrivAerNet Center Plane & \texttt{drivaernet\_3d\_2d\_centerplane} & 3d$\to$2d & pressure, velocity & CC BY-NC 4.0 \\
 & DrivAerNet Surface & \texttt{drivaernet\_3d\_2d\_surface} & 3d$\to$2d & pressure & CC BY-NC 4.0 \\
\bottomrule
\end{tabular}

\end{table*}

\subsection{Tasks and Model Families}
\label{sec:method:models}

The task taxonomy operationalizes the generative-AI design-optimization framework of Kang~\cite{kang2025scenarios}, which organizes engineering AI by data dimensionality across 1D, 2D, and 3D and by model role across generation, prediction, and optimization. \benchname{} realizes the generative and predictive axes as a ranked benchmark. Design optimization searches for new designs rather than scoring model outputs and lies outside the evaluation scope. This yields seven tasks. The 1D domain contributes scalar prediction. The 2D domain contributes image generation, scalar prediction, and field prediction. The 3D domain contributes geometry generation, scalar prediction, and field prediction.

\benchname{} evaluates 66 publicly ranked models, together with the official AB-UPT described below, on nine source datasets, expanded into the 28 task-specific configurations of Table~\ref{tab:datasets} and each swept over four controlled data scales from S to XL. Coverage is uniform within a task with one exception. In 3D field prediction, four standalone models, GeoFLARE, GeoTransolver, DoMINO, and the official AB-UPT, are evaluated on DrivAerML alone, the automotive-aerodynamics dataset they were built for, so their pooled scores rest on narrower coverage than the full-coverage models and are read per dataset in Section~\ref{sec:results:pertask}. AB-UPT is further reported under a research license that permits evaluation use but bars public deployment, so it is ranked on the DrivAerML boards in this paper and withheld from the public leaderboard. Its scores are our own measurements, obtained by training and evaluating the authors' released implementation under the same procedure applied to every other model, rather than values quoted from its source paper. Each task hosts a family of models implemented as described in their source papers, with occasional capacity-reduced variants for the small-data regime documented as distinct entries.

Generation spans GAN~\cite{goodfellow2014gan}, VAE~\cite{kingma2014vae}, and diffusion~\cite{ho2020ddpm} models in 2D, and SDF~\cite{park2019deepsdf}, GAN~\cite{wu20163dgan}, flow~\cite{yang2019pointflow}, score-based~\cite{cai2020shapegf}, autoencoder~\cite{groueix2018atlasnet}, and diffusion~\cite{luo2021diffusionpc} models in 3D. Scalar regression uses point-cloud backbones~\cite{qi2017pointnet2} in 3D, CNN and vision-transformer (ViT) image backbones~\cite{he2016resnet,touvron2021deit} in 2D, and tabular and time-series models in 1D. The 1D tabular models are an MLP, FT-Transformer~\cite{gorishniy2021ftt}, NODE~\cite{popov2020node}, TabNet~\cite{arik2021tabnet}, TabPFN~\cite{hollmann2025tabpfn}, gradient-boosted trees~\cite{chen2016xgboost,ke2017lightgbm}, Gaussian processes~\cite{rasmussen2006gp}, and ridge regression~\cite{hoerl1970ridge}, and the 1D time-series models target remaining-useful-life prediction~\cite{zheng2017lstm,zhang2022dast}. Field prediction uses neural operators and geometry transformers, including Transolver~\cite{wu2024transolver}, LinearNO~\cite{linearno2026}, Geo-FNO~\cite{li2023geofno}, GeoTransolver~\cite{adams2025geotransolver}, GeoFLARE~\cite{akhare2026geoflare,puri2025flare}, DoMINO~\cite{nvidia2025domino}, and
AB-UPT~\cite{alkin2025abupt}, together with point- and graph-based backbones such as PointNet~\cite{qi2017pointnet} and RegDGCNN~\cite{wang2019dgcnn} in 3D, and U-Net~\cite{ronneberger2015unet}, FPN~\cite{lin2017fpn}, and dense-prediction transformer~\cite{ranftl2021dpt} architectures in 2D.

The predictive family is predominantly Computer-Aided Engineering (CAE) emulators for structural, aerodynamic, and porous-flow fields and their derived quantities. It also includes a small set of adjacent tasks that share the same input-to-output regression over quantities of interest (QoIs), the same metric suite, and the same evaluation procedure. These are a monocular-depth reference task, a computer-vision problem included for cross-domain reference rather than a CAE surrogate, and prognostic remaining-useful-life estimation. Generative and predictive families share the same procedure and leaderboard but are ranked within their own tasks and metric suites, not head-to-head. The complete per-task inventories, with parameter counts and references, appear in Table~\ref{tab:models}.

\begin{table*}[t]
  \centering
  \footnotesize
  \caption[Models evaluated in \benchname{}, grouped by task]%
  {Models evaluated in \benchname{}, grouped by task. Parameter counts
  are read from the released results, and \texttt{n/a} marks the estimators for
  which the benchmark records none, namely the tree, kernel, and in-context
  tabular baselines. The 3D field inventory additionally includes the official
  AB-UPT, reported on DrivAerML only.}
  \label{tab:models}
  \begin{tabular}{@{}l l r l@{\hspace{2.2em}}l l r l@{}}
\toprule
Model & Category & Params (M) & Reference & Model & Category & Params (M) & Reference \\
\midrule
\multicolumn{4}{@{}l}{\textbf{1D scalar prediction}} & \multicolumn{4}{@{}l}{\textbf{2D field prediction}} \\
BiLSTM & RNN & 0.15 & 2018~\cite{wang2018bilstm} & Attention U-Net & U-Net & 7.98 & MIDL 2018~\cite{oktay2018attentionunet} \\
CNN-LSTM & Hybrid & 0.04 & — & DPT-Hybrid & DPT & 122.38 & ICCV 2021~\cite{ranftl2021dpt} \\
DAST & Transformer & 0.07 & 2022~\cite{zhang2022dast} & FPN (ResNet-18) & FPN & 15.56 & CVPR 2017~\cite{lin2017fpn,kirillov2019panopticfpn,he2016resnet} \\
DCNN & CNN & 0.02 & 2018~\cite{li2018dcnn} & GLPN & DPT & 61.22 & 2022~\cite{kim2022glpn} \\
FT-Transformer & Tabular-DL & 0.10 & NeurIPS 2021~\cite{gorishniy2021ftt} & ResNet-UNet & U-Net & 14.41 & MICCAI 2015~\cite{ronneberger2015unet,he2016resnet} \\
Gaussian Process & Classical & n/a & 2006~\cite{rasmussen2006gp} & SegFormer-B0 & Transformer & 3.71 & NeurIPS 2021~\cite{xie2021segformer} \\
LightGBM & Tree & n/a & NeurIPS 2017~\cite{ke2017lightgbm} & U-Net & U-Net & 7.85 & MICCAI 2015~\cite{ronneberger2015unet} \\
LSTM & RNN & 0.06 & 2017~\cite{zheng2017lstm} & U-Net++ & U-Net & 9.16 & DLMIA 2018~\cite{zhou2018unetpp} \\
MLP & MLP & 0.04 & — & \multicolumn{4}{@{}l}{\textbf{3D geometry generation}} \\
NODE & Tabular-DL & 0.01 & ICLR 2020~\cite{popov2020node} & 3D-GAN & GAN & 74.98 & NeurIPS 2016~\cite{wu20163dgan} \\
Random Forest & Classical & n/a & 2001~\cite{breiman2001rf} & AtlasNet & AtlasNet & 28.77 & CVPR 2018~\cite{groueix2018atlasnet} \\
Ridge & Classical & n/a & 1970~\cite{hoerl1970ridge} & DeepSDF & SDF & 2.15 & CVPR 2019~\cite{park2019deepsdf} \\
TabNet & Tabular-DL & 0.01 & AAAI 2021~\cite{arik2021tabnet} & Diffusion3D & Diffusion & 3.88 & CVPR 2021~\cite{luo2021diffusionpc} \\
TabPFN & Foundation & n/a & Nature 2025~\cite{hollmann2025tabpfn} & PointFlow & Flow & 1.43 & ICCV 2019~\cite{yang2019pointflow} \\
TCN & CNN & 0.03 & 2018~\cite{bai2018tcn} & ShapeGF & Score & 4.86 & ECCV 2020~\cite{cai2020shapegf} \\
XGBoost & Tree & n/a & KDD 2016~\cite{chen2016xgboost} & \multicolumn{4}{@{}l}{\textbf{3D scalar prediction}} \\
\multicolumn{4}{@{}l}{\textbf{2D image generation}} & DGCNN & Graph & 1.80 & ACM TOG 2019~\cite{wang2019dgcnn} \\
DCGAN & GAN & 23.95 & ICLR 2016~\cite{radford2016dcgan} & PCT & Transformer & 1.24 & CVM 2021~\cite{guo2021pct} \\
DDPM & Diffusion & 71.44 & NeurIPS 2020~\cite{ho2020ddpm} & PCT-Small & Transformer & 0.36 & CVM 2021~\cite{guo2021pct} \\
GAN (Basic) & GAN & 76.37 & NeurIPS 2014~\cite{goodfellow2014gan} & Point Transformer & Transformer & 12.15 & ICCV 2021~\cite{zhao2021pointtransformer} \\
LSGAN & GAN & 39.09 & ICCV 2017~\cite{mao2017lsgan} & Point Transformer-Small & Transformer & 0.83 & ICCV 2021~\cite{zhao2021pointtransformer} \\
R1GAN & GAN & 23.95 & ICML 2018~\cite{mescheder2018r1} & PointMLP & MLP & 3.14 & ICLR 2022~\cite{ma2022pointmlp} \\
VAE (Basic) & VAE & 50.72 & ICLR 2014~\cite{kingma2014vae} & PointMLP-Elite & MLP & 0.48 & ICLR 2022~\cite{ma2022pointmlp} \\
VQVAE & VAE & 4.89 & NeurIPS 2017~\cite{oord2017vqvae} & PointNet & PointNet & 0.08 & CVPR 2017~\cite{qi2017pointnet} \\
WGAN-CP & GAN & 23.95 & ICML 2017~\cite{arjovsky2017wgan} & PointNet++ & PointNet & 1.47 & NeurIPS 2017~\cite{qi2017pointnet2} \\
WGAN-GP & GAN & 23.95 & NeurIPS 2017~\cite{gulrajani2017wgangp} & PointNet++ Lite & PointNet & 0.27 & adapted from~\cite{qi2017pointnet2} \\
\multicolumn{4}{@{}l}{\textbf{2D scalar prediction}} & \multicolumn{4}{@{}l}{\textbf{3D field prediction}} \\
ConvNeXt-Tiny & CNN & 28.02 & CVPR 2022~\cite{liu2022convnext} & AB-UPT & UPT & 8.75 & TMLR 2025~\cite{alkin2025abupt} \\
DenseNet-121 & CNN & 7.22 & CVPR 2017~\cite{huang2017densenet} & DoMINO & Neural Operator & 24.31 & 2025~\cite{nvidia2025domino} \\
EfficientNet-B0 & CNN & 4.34 & ICML 2019~\cite{tan2019efficientnet} & GeoFLARE & Transformer & 26.41 & 2026~\cite{akhare2026geoflare,puri2025flare,adams2025geotransolver} \\
ResNet-18 & CNN & 11.31 & CVPR 2016~\cite{he2016resnet} & GeoFNO & Neural Operator & 1.55 & JMLR 2023~\cite{li2023geofno} \\
ResNet-34 & CNN & 21.42 & CVPR 2016~\cite{he2016resnet} & GeoTransolver & Transformer & 29.49 & 2025~\cite{adams2025geotransolver} \\
SimpleCNN & CNN & 0.42 & — & LinearNO & Neural Operator & 3.33 & AAAI 2026~\cite{linearno2026} \\
ViT-Tiny & Transformer & 5.55 & ICML 2021~\cite{touvron2021deit} & LinearNO-Big & Neural Operator & 11.05 & AAAI 2026~\cite{linearno2026} \\
 &  &  &  & PointNet & PointNet & 3.67 & CVPR 2017~\cite{qi2017pointnet} \\
 &  &  &  & RegDGCNN & Graph & 1.44 & ACM TOG 2019~\cite{wang2019dgcnn} \\
 &  &  &  & Transolver & Transformer & 3.78 & ICML 2024~\cite{wu2024transolver} \\
 &  &  &  & Transolver++ & Transformer & 1.74 & ICML 2025~\cite{luo2025transolverpp} \\
\bottomrule
\end{tabular}

\end{table*}

\subsection{Unified Pipeline}
\label{sec:method:pipeline}

The pipeline has three steps. \textbf{Step~A} trains a model on a fixed number of samples and then generates standardized outputs for inference. \textbf{Step~B} loads the real and predicted samples, computes the task's metrics, and writes a single per-model result file. \textbf{Step~C} aggregates these files and ranks the models, as specified in Section~\ref{sec:method:benchrank}. Training conventions are fixed per domain and task in Table~\ref{tab:conventions} so that all models in a task are trained under identical budgets. This fixes the shared procedure of data splits, resolution, epoch and batch budget, early stopping, and normalization, but not each model's own optimizer, learning rate, and internal hyperparameters, which follow its source paper. A ranking difference is therefore attributable to the model as published, meaning its architecture and recommended configuration, and in particular is not explained by a larger-data or longer-training advantage, rather than to a controlled single-hyperparameter ablation. The specific values follow
three practical constraints. Resolutions of $8192$-point clouds and $128{\times}128$ images are each domain's de-facto standard and, deliberately, the common denominator at which architectures spanning roughly five years, from established backbones to the most recent operators, run on their published configurations without structural surgery, which keeps the comparison about the models rather than about bespoke re-tuning. Batch sizes are the largest that fit GPU memory at those resolutions, 8 for the memory-heavy 3D point sets and 16 in 2D, and the higher generation budget of 1000 epochs against 500 reflects the slower convergence of adversarial, diffusion, and flow training, with early stopping trimming both in practice. The data-scale buckets are not an arbitrary cap. They mirror the reality that in engineering practice the number of accessible high-fidelity samples is small, therefore, S--XL span the regime engineers actually operate in, with 2D generation shifted upward to 50--500 only because FID is unstable and upward-biased at small sample sizes~\cite{chong2020fid}. Early stopping monitors a domain-appropriate validation metric, namely FID for 2D generation, a Fr\'echet PointNet distance (FPD) for 3D generation, and validation MSE for all prediction
tasks, checked every ten epochs, with patience of 50 checks for generation or 25 checks for prediction and a minimum of 50 or 20 epochs respectively. Inference always uses the best-validation checkpoint rather than the final epoch, and produces up to 500 samples, capped at the test-split size. Generative models that lack a native sampling prior, most notably the auto-decoder SDF model that learns one latent code per training shape rather than a generator network, are sampled for generation by fitting a Gaussian mixture to their learned training latents and decoding draws from it, so their generative quality depends on how well that latent density can be estimated from the available shapes. Training and inference wall-clock times are recorded as cost metrics.

\begin{table}[t]
  \centering
  \footnotesize
  \caption[Fixed training conventions per domain and task]%
  {Fixed training conventions per domain and task. Sizes define the
  S/M/L/XL data-scale buckets used by the leaderboard.}
  \label{tab:conventions}
  \begin{tabular}{@{}l c c c l@{}}
    \toprule
    Domain/task & Resolution & Epochs & Batch & Sizes \\
    \midrule
    1D scalar         & n/a              & 1000 & 16 & 20/50/100/200 \\
    2D generation     & $128{\times}128$ & 1000 & 16 & 50/100/200/500 \\
    2D scalar/field   & $128{\times}128$ & 500  & 16 & 20/50/100/200 \\
    3D generation     & 8192 pts         & 1000 & 8  & 20/50/100/200 \\
    3D scalar/field   & 8192 pts         & 500  & 8  & 20/50/100/200 \\
    \bottomrule
  \end{tabular}
\end{table}

Each model run writes one result file. A new model is added by committing that single file, which avoids merge conflicts between contributors. The leaderboard is a deterministic function of the committed results that any user can regenerate. Non-gradient estimators such as TabPFN and gradient-boosted trees bypass the gradient trainer through a thin scikit-learn-style adapter but reuse the identical inference and scoring path, hence the comparison remains fair.

Every model is trained at a fixed seed, and cells are additionally rerun at two further seeds. A leaderboard is a pairwise comparison, so the reporting basis is fixed per task and never per model, and every model on one board is measured the same way. Ranking averaged models against un-averaged ones would strip the run-to-run luck from some competitors and not others, which is the one asymmetry a max-over-models ranking cannot absorb. A task is therefore reported as the per-cell mean over the three runs only where every one of its cells carries the complete rerun set, and as the single baseline run otherwise. Five of the seven tasks meet that condition, namely 1D scalar, 2D generation, 2D scalar, 2D field, and 3D scalar prediction. The two that do not are 3D generation, where four of its $84$ cells are short a rerun, and 3D field prediction, where three of the four standalone models named in Section~\ref{sec:results:pertask} are reported from a single run, leaving $236$ of its $1526$ cells unreplicated. \ref{app:robustness} reports how far the rankings move with the training run, including what changes when every task is placed on the mean of its repeated runs.

\subsection{Metric Suite}
\label{sec:method:metrics}

Metrics are organized into three families and selected according to the task, as summarized in Table~\ref{tab:metrics}. The geometric-fidelity family evaluates generated images and point clouds using distributional, perceptual, and geometric measures. For 2D generation, the metrics include FID~\cite{heusel2017fid}, Inception Score~\cite{salimans2016is}, precision, recall, density, and coverage~\cite{naeem2020prdc}, LPIPS~\cite{zhang2018lpips}, and MS-SSIM~\cite{wang2004ssim}. For 3D generation, the metrics include Fr\'echet PointNet distance and the set-level Chamfer-based measures MMD, COV, and 1-NNA~\cite{achlioptas2018pointcloud,yang2019pointflow}, together with precision, recall, density, and coverage. The prediction-accuracy family evaluates physical fields and scalar outputs using MAE, RMSE, MAPE, $R^2$, and relative $L_2$ error, supplemented by task-specific measures. Vector and modal-field tasks use the Modal Assurance Criterion (MAC)~\cite{allemang1982mac}, sign agreement, and extremal agreement. Depth-regression tasks use AbsRel, sqRel, and $\delta{<}1.25$~\cite{eigen2014depth}. Scalar-prediction tasks additionally use maximum absolute error and the Pearson and Spearman correlation coefficients. The computational-cost family reports parameter count, training time, and inference time.

Two of these metric families assess engineering validity rather than relying exclusively on the pixel-wise or point-wise mean errors commonly used in ML benchmarks. \benchname{} extends conventional leaderboard evaluation with five ranked engineering-validity metrics. Four metrics, \emph{sign agreement}, \emph{extremal agreement}, \emph{manifold-$\Delta$}, and \emph{uniformity-$\Delta$}, are introduced in this work. The fifth, the Modal Assurance Criterion (MAC) \cite{allemang1982mac}, is adapted from structural dynamics. All five metrics are ranked alongside mean error because they quantify failure modes that standard error and distributional metrics do not explicitly isolate.

For field prediction, a low MAE does not guarantee physically correct behavior. A model may predict the wrong loading direction or place stress or pressure extrema in incorrect regions. Sign agreement measures sign consistency between the predicted and reference fields, while extremal agreement measures the spatial agreement between their extrema. The sign and scale-invariant MAC measures the alignment of mode shapes and vector-field patterns up to a global sign and amplitude, thereby separating structural-pattern agreement from magnitude error.

For geometry generation, manifold-$\Delta$ measures local manifold validity, whereas uniformity-$\Delta$ measures the uniformity of point sampling over the generated geometry. These metrics distinguish two different geometric failures, namely local non-manifold structure and nonuniform point concentration such as clustering along seams. Standard distributional metrics, including FID, FPD, precision, recall, density, and coverage, do not explicitly isolate either failure. Consequently, a generator may obtain favorable distributional scores while producing point clouds that are unsuitable for downstream meshing or simulation.

Such engineering-specific measures are generally absent from general-purpose ML leaderboards or are treated as secondary to a single fidelity metric. \benchname{} instead treats engineering validity as a first-class ranking criterion because a prediction or generated design with low conventional error but invalid physics or geometry remains unusable. \ref{app:metrics} provides formal definitions of these metrics and of the standard metrics listed above.

Consistent with this design principle, every displayed quality metric is included in the ranking, and no quality metric is reported for display only. Each task is therefore reported under the two leaderboard views defined in Section~\ref{sec:method:benchrank}. The quality view carries the computational-cost metrics for reference at zero weight, so the ranking is decided by quality alone. The efficiency view promotes those same cost metrics to ranked metrics beside the quality ones, so a model is scored at once on what it delivers and on what it costs. Those cost metrics are the subset $\mathcal{E}$ of Algorithm~\ref{alg:benchrank}. Each view therefore ranks models using exactly the criteria assigned nonzero weight in it.

\begin{table}[t]
  \centering
  \footnotesize
  \caption[Metric suite by task family]%
  {Metric suite by task family, where $\uparrow$ and $\downarrow$ mark
  higher-is-better and lower-is-better. The scalar-prediction row applies to the
  1D, 2D, and 3D scalar tasks. All tasks additionally report parameters and
  train and infer time. The generation metrics LPIPS, MS-SSIM, and PSNR follow
  the direction conventions defined in \ref{app:metrics}.}
  \label{tab:metrics}
  \begin{tabular}{@{}l p{0.62\columnwidth}@{}}
    \toprule
    Task family & Metrics \\
    \midrule
    2D generation & IS\,$\uparrow$, FID\,$\downarrow$, LPIPS\,$\uparrow$, PSNR\,$\uparrow$, MS-SSIM\,$\downarrow$, Precision\,$\uparrow$, Recall\,$\uparrow$, Density\,$\uparrow$, Coverage\,$\uparrow$ \\
    \addlinespace
    3D generation & FPD\,$\downarrow$, MV-FID\,$\downarrow$, MMD-CD\,$\downarrow$, COV-CD\,$\uparrow$, 1-NNA-CD\,$\downarrow$, MS-SSIM\,$\downarrow$, Precision\,$\uparrow$, Recall\,$\uparrow$, Density\,$\uparrow$, Coverage\,$\uparrow$, Manifold-$\Delta$\,$\downarrow$, Uniformity-$\Delta$\,$\downarrow$ \\
    \addlinespace
    Scalar prediction & MAE\,$\downarrow$, RMSE\,$\downarrow$, MAPE\,$\downarrow$, $R^2$\,$\uparrow$, Rel-$L_2$\,$\downarrow$, MaxAE\,$\downarrow$, Pearson\,$\uparrow$, Spearman\,$\uparrow$ \\
    \addlinespace
    Field prediction & MAE\,$\downarrow$, RMSE\,$\downarrow$, MAPE\,$\downarrow$, $R^2$\,$\uparrow$, Rel-$L_2$\,$\downarrow$, and for 3D also MAC\,$\uparrow$, Sign Agree\,$\uparrow$, Extremal Agree\,$\uparrow$, for 2D also PSNR\,$\uparrow$, SSIM\,$\uparrow$, and for depth AbsRel\,$\downarrow$, sqRel\,$\downarrow$, $\delta{<}1.25$\,$\uparrow$ \\
    \bottomrule
  \end{tabular}
\end{table}

\subsection{BenchRank: Debiased Graph-Based Ranking}
\label{sec:method:benchrank}

A practitioner-facing leaderboard must reduce a task's many metrics to a single ranking, but the obvious reductions all mislead. Ranking by one headline metric such as FID alone or $R^2$ alone discards the rest of the suite, including the very field- and shape-validity metrics that distinguish a usable prediction from a metrically-close but physically wrong one. Averaging a normalized score instead double-counts correlated metrics, since MAE and RMSE move together, and lets a single large-scale metric dominate a naive mean. We are aware of no engineering benchmark that supplies a multi-metric ranking correcting for this. We therefore introduce \textbf{BenchRank}, a graph-based ranking algorithm designed for this setting. It adopts two established primitives as building blocks, Spearman-correlation debiasing of the metric weights and PageRank centrality for scoring. What BenchRank contributes is the head-to-head dominance graph it constructs over the full metric suite, the viability gate that removes degenerate models, and the clipped geometric-mean aggregation that makes cross-dataset scores comparable. Models are compared pairwise across the whole metric suite at once. For each ordered pair of models we accumulate a signed magnitude over the metrics, adding a term wherever the first model beats the second by more than a per-metric threshold and subtracting a term wherever it is beaten by more than that threshold, with each term growing as $\log(1{+}z^{2})$ in the standardized gap $z$, or as $\log(1{+}z)$ for cost metrics. Whenever this net margin is positive we add a single directed edge from the weaker to the stronger model as an endorsement, weighted by the net margin and detailed in Algorithm~\ref{alg:benchrank}. Each metric's threshold is the $1{-}\tau$ quantile of that metric's observed pairwise gaps at $\tau{=}0.6$, so that roughly a fraction $\tau$ of pairs count as decisive on it while the remainder are suppressed as ties, which leaves the dominance graph neither fully connected nor disconnected. The threshold is clipped to between $0.05$ and $5$ median absolute deviations of that metric, so that a metric whose gaps are heavy-tailed relative to its dispersion cannot set its own threshold from outliers alone. Metric weights are debiased by redundancy. Each metric's weight is $w = 1/(1+\alpha\,\rho)$, where $\rho$ is its summed Spearman correlation with the other metrics at $\alpha{=}1$, so mutually redundant metrics are down-weighted. A model's score is its PageRank~\cite{page1999pagerank} centrality in this weighted endorsement graph, which is high when the model is preferred by many strong competitors across the metric suite, computed with damping $0.85$ via \texttt{networkx}~\cite{hagberg2008networkx} and scaled to a $[0,100]$ Total Score.

Two robustness mechanisms complete the ranking. A viability gate excludes degenerate models. A model that is a median-absolute-deviation outlier at $k{=}3$ on all of a task's gated metrics is removed from the graph and assigned a floor score below all viable models. Mode collapse itself needs no dedicated rule, because a collapsed generator scores poorly on the coverage and distributional metrics and is demoted to the lower ranks by the score on its own. We deliberately apply no analogous gate to the regression tasks. A cell in which every model scores $R^2{<}0$ is not noise to be removed but decision-relevant information, since it tells a practitioner the task is unlearnable at that scale, so we retain and rank it and flag it as indicative. This signal is reflected in the BenchRank score rather than gated out of it. Within a viable cell, a single mean-collapsing model is instead demoted by the rank-correlation metrics Pearson and Spearman, which a constant predictor cannot satisfy, as discussed in Section~\ref{sec:results:scalar}. A dedicated regression viability gate remains a possible future refinement. Across evaluation cells, one per dataset and quantity channel, a model's Total Score is aggregated by a clipped geometric mean with floor $0.01$ rather than an arithmetic mean, so a model that collapses on any single cell is penalized more heavily than one that is uniformly mediocre. A model evaluated on only a subset of a task's cells aggregates over that subset alone, so its pooled score is not directly commensurable with a full-coverage model's and must be read per dataset, as detailed in Section~\ref{sec:results:pertask}. Ranking ties break deterministically by model name, making the leaderboard reproducible regardless of input order. Leaderboards are produced per data-scale bucket from S to XL in both the quality and the efficiency view. Algorithm~\ref{alg:benchrank} states the complete procedure, from dominance-graph construction through the viability gate and the cross-cell geometric mean.

A final clarification concerns how the figures report ranking versus raw metrics. BenchRank's total score is a relative centrality re-normalized within each task and scale bucket so that the top model scores $100$, while the bottom of the interval rises with the share of model pairs the metric suite separates, from $20$ when it separates none to $40$ when it separates all, and lies between $32$ and $40$ across the leaderboards reported here. The score is therefore deliberately not comparable across scales or tasks and carries no physical unit. The trajectory and Pareto figures of Section~\ref{sec:results} therefore plot the raw primary metric, namely $R^2$, FID, or FPD, whose absolute magnitude and cross-scale trend the score cannot express. The score is thus the selection layer that says which model to pick within a bucket, and the raw metrics are the evidence layer that says how good each model is in physical units. We report both rather than collapsing one into the other.

\begin{algorithm}[t]
\caption{\textbf{BenchRank}: debiased graph-based ranking for one task at a fixed data scale.}
\label{alg:benchrank}
\begin{algorithmic}[1]
\Require per-cell metric matrices $\{M^{c}\}_{c\in\mathcal{C}}$, one evaluation cell per
  (dataset, quantity channel), each $M^{c}\!\in\!\mathbb{R}^{n\times m}$ ($n$ models,
  $m$ metrics), with per-metric direction ($\uparrow$/$\downarrow$); cost-metric subset
  $\mathcal{E}$ and margin map $g(z){=}\log(1{+}z^{2})$ on the quality metrics
  $j \notin \mathcal{E}$, $\log(1{+}z)$ on $\mathcal{E}$; gated-metric subset $G$; $\mathrm{MAD}_j$ the
  normal-consistency-scaled ($1.4826{\times}$) median absolute deviation of metric $j$;
  redundancy strength $\alpha$, decisive-gap fraction $\tau$, threshold bounds
  $\gamma_{\min}{<}\gamma_{\max}$ in units of $\mathrm{MAD}_j$, PageRank damping
  $\delta$, gate factor $k$, geometric-mean floor $\varepsilon$
\Ensure Total Score $s\in[0,100]^{n}$ and the induced ranking
\For{each evaluation cell $c \in \mathcal{C}$}
  \State $V \gets$ all models except those worse than $\mathrm{median}_j\!+\!k\,\mathrm{MAD}_j$ on every $j \in G$ \Comment{viability gate}
  \State restrict $M^{c}$ to $V$ and negate every $\downarrow$ metric, so larger is better
  \Statex \quad\textit{-- debiased weights and thresholds --}
  \For{each metric $j$}
    \State $\rho_j \gets \textstyle\sum_{l \neq j} \lvert \mathrm{Spearman}(M^{c}_{:,j}, M^{c}_{:,l}) \rvert$ \Comment{summed rank-correlation}
    \State $w_j \gets 1 / (1 + \alpha\,\rho_j)$ \Comment{redundant metrics down-weighted}
    \State $\theta_j \gets \mathrm{clip}\big(\mathrm{quantile}_{1-\tau}\{\,|M^{c}_{aj}{-}M^{c}_{bj}| : a{<}b\,\},\ [\gamma_{\min},\gamma_{\max}]\,\mathrm{MAD}_j\big)$ \Comment{$\approx\!\tau$ of pairs exceed $\theta_j$}
  \EndFor
  \State normalize the weights so that $\sum_j w_j = 1$
  \Statex \quad\textit{-- weighted endorsement graph (net pairwise dominance) --}
  \State $E \gets \varnothing$
  \For{each unordered pair $\{a,b\} \subseteq V$}
    \State $\Delta_j \gets M^{c}_{aj} - M^{c}_{bj}$ for every metric $j$
    \State $\displaystyle \nu \gets \!\!\sum_{j:\,|\Delta_j|>\theta_j}\!\! \mathrm{sign}(\Delta_j)\, w_j\, g\!\big(|\Delta_j|/\theta_j\big)$ \Comment{net margin of $a$ over $b$}
    \If{$\nu \neq 0$}
      \State add edge loser $\to$ winner, weight $|\nu|$ \Comment{$a$ wins iff $\nu{>}0$}
    \EndIf
  \EndFor
  \Statex \quad\textit{-- PageRank centrality, then rescale --}
  \State $p \gets \mathrm{PageRank}(E, \delta)$;\quad $\hat d \gets |E| / \big(|V|(|V|{-}1)\big)$ \Comment{realized density, $\hat d \le 1/2$}
  \State affinely map $p$ over $V$ onto $[\,20{+}40\hat d,\ 100\,]$, giving $s^{c}$ \Comment{top model scores $100$}
  \State assign each non-viable model a floor below $\min_{i \in V} s^{c}_{i}$
\EndFor
\Statex \textit{-- cross-cell aggregation --}
\State for each model $i$: $\mathcal{C}_i \gets \{\, c \in \mathcal{C} : i \text{ evaluated on } c \,\}$ \Comment{each model's own coverage}
\State $s_i \gets \big( \textstyle\prod_{c \in \mathcal{C}_i} \max(s^{c}_{i}, \varepsilon) \big)^{1/|\mathcal{C}_i|}$ \Comment{clipped geo.\ mean; partial coverage not commensurable}
\State \Return $s$ and the ranking, breaking ties by model name \Comment{deterministic, input-order independent}
\end{algorithmic}
\end{algorithm}

\section{Results and Discussion}
\label{sec:results}

We evaluate the full model set under the procedure of Section~\ref{sec:method}.
Wall-clock training and inference times are reported on a common NVIDIA RTX PRO
6000 with 96\,GB of memory, with each model, dataset, and scale cell trained and
evaluated on a single GPU. Running the full suite on a single device calls for a
high-memory card, because DoMINO's footprint exceeds the 80\,GB of an A100, and
the 96\,GB here accommodates every model with room to spare. The central
empirical finding is twofold. Within a task, the best model often changes
with the data scale, and across the benchmark no single architecture is
best everywhere. This holds not merely because the tasks differ, but because
rankings reorder with scale and dataset even among models that compete
head-to-head, and both effects are obscured by conventional single-task,
single-scale evaluation.

\subsection{No Universal Winner}
\label{sec:results:winners}

Table~\ref{tab:winners} reports the top-ranked model for each task at each data
scale. No single architecture wins across tasks, and the top model changes as the
training-set size grows in six of the seven tasks, 3D field prediction being the
only exception. The 1D scalar task is reported as two rows,
tabular regression on Concrete and Airfoil and time-series RUL on CMAPSS, because
these span disjoint model sets, so a single pooled ranking would compare models
that never competed head-to-head. A practitioner cannot, therefore, read a single
``best model'' off the benchmark. The right choice is conditioned on the task
and the available data scale, which is precisely the decision support the
leaderboard is designed to provide.

\begin{table*}[t]
  \centering\footnotesize
  \caption[Top-ranked model per task and data scale]%
  {Top-ranked model per task and data scale under the BenchRank quality
  view. The 1D scalar task is reported as two rows, one per model pool, namely
  tabular and time-series RUL.}
  \label{tab:winners}
  \resizebox{\textwidth}{!}{\begin{tabular}{@{}l l l l l@{}}
\toprule
Task & S & M & L & XL \\
\midrule
1D scalar: tabular (Concrete, Airfoil) & TabPFN & TabPFN & TabPFN & NODE \\
1D scalar: time-series RUL (CMAPSS) & CNN-LSTM & TCN & TCN & DAST \\
2D image generation & VAE (Basic) & DDPM & DDPM & DDPM \\
2D scalar prediction & SimpleCNN & SimpleCNN & ConvNeXt-Tiny & ConvNeXt-Tiny \\
2D field prediction & U-Net++ & U-Net++ & U-Net++ & FPN (ResNet-18) \\
3D geometry generation & DeepSDF & DeepSDF & DeepSDF & PointFlow \\
3D scalar prediction & Point Transformer-Small & DGCNN & Point Transformer-Small & Point Transformer-Small \\
3D field prediction & GeoFLARE & GeoFLARE & GeoFLARE & GeoFLARE \\
\bottomrule
\end{tabular}
}
  \\[3pt]
  \begin{minipage}{\textwidth}\footnotesize\raggedright
  \emph{Note.} The 3D-field winner GeoFLARE is a standalone model
  evaluated on DrivAerML only, so its pooled rank reflects narrower coverage than
  the full-coverage baselines and must be read per dataset, as detailed in
  Section~\ref{sec:results:pertask}. AB-UPT is additionally reported under
  a research license and excluded from the public leaderboard.
  \end{minipage}
\end{table*}

\subsection{Data-Scale Crossovers and Small-Data Behavior}
\label{sec:results:scaling}

\begin{figure*}[t]
  \centering
  \begin{minipage}{0.49\textwidth}\centering
    \includegraphics[width=\linewidth]{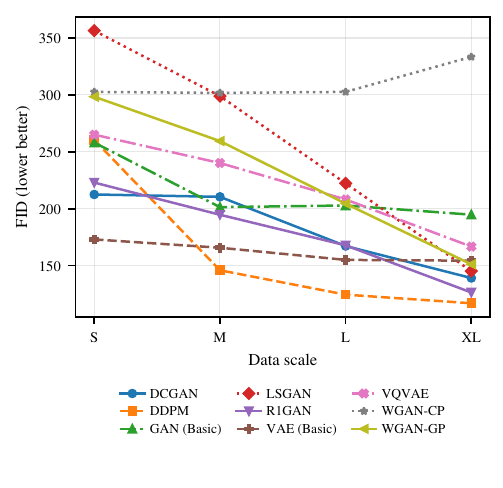}\\[-1pt]
    {\footnotesize(\emph{a}) 2D generation (FID$\downarrow$)}
  \end{minipage}\hfill
  \begin{minipage}{0.49\textwidth}\centering
    \includegraphics[width=\linewidth]{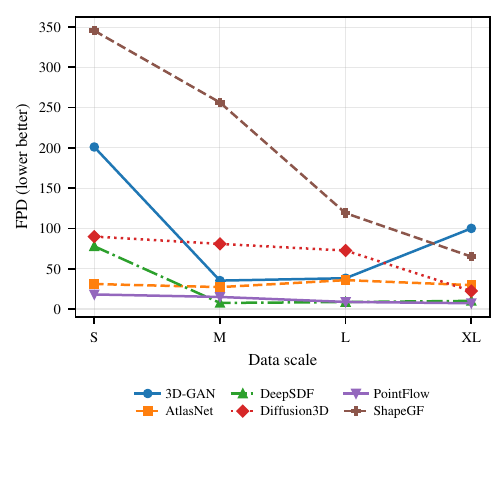}\\[-1pt]
    {\footnotesize(\emph{b}) 3D generation (FPD$\downarrow$)}
  \end{minipage}
  \caption[Primary metric versus data scale for generation]%
  {Primary metric versus data scale, one line per model, for (\emph{a})~2D
  generation, scored by FID, and (\emph{b})~3D generation, scored by FPD.
  Prediction-task trajectories are in Figure~\ref{fig:trajectory_pred}.}
  \label{fig:trajectory}
\end{figure*}

\begin{figure*}[t]
  \centering
  \begin{minipage}{0.49\textwidth}\centering
    \includegraphics[width=\linewidth]{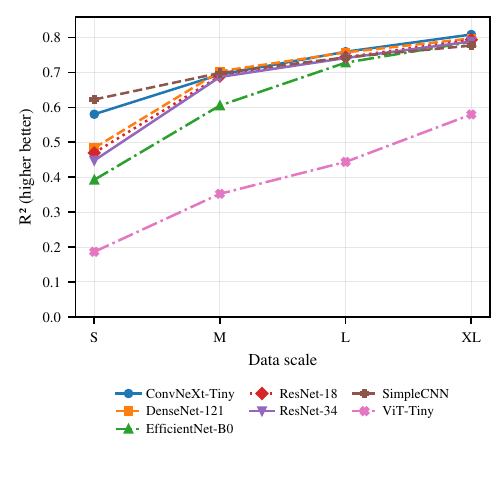}\\[-1pt]
    {\footnotesize(\emph{a}) 2D scalar prediction ($R^2\uparrow$)}
  \end{minipage}\hfill
  \begin{minipage}{0.49\textwidth}\centering
    \includegraphics[width=\linewidth]{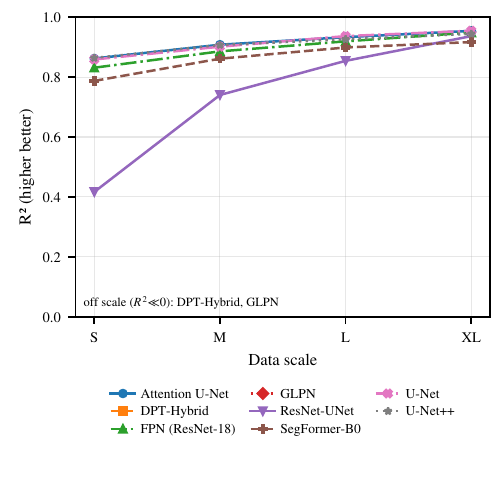}\\[-1pt]
    {\footnotesize(\emph{b}) 2D field prediction ($R^2\uparrow$)}
  \end{minipage}\\[4pt]
  \begin{minipage}{0.49\textwidth}\centering
    \includegraphics[width=\linewidth]{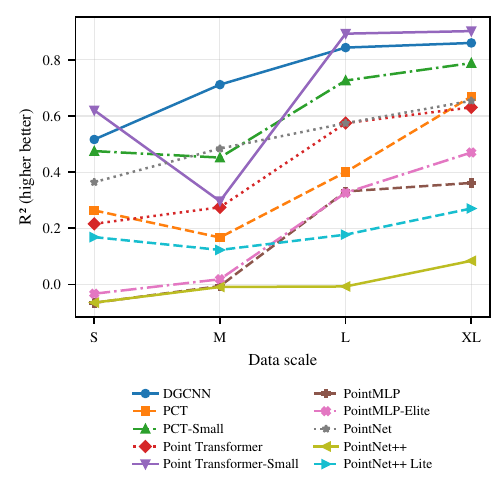}\\[-1pt]
    {\footnotesize(\emph{c}) 3D scalar prediction ($R^2\uparrow$)}
  \end{minipage}\hfill
  \begin{minipage}{0.49\textwidth}\centering
    \includegraphics[width=\linewidth]{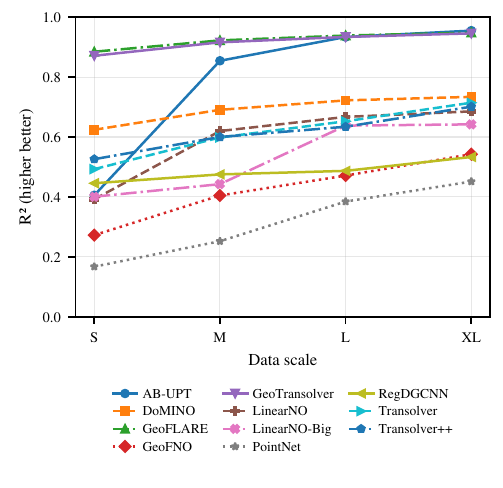}\\[-1pt]
    {\footnotesize(\emph{d}) 3D field prediction ($R^2\uparrow$)}
  \end{minipage}
  \caption[Primary accuracy versus data scale for prediction]%
  {Primary accuracy $R^2$ versus data scale from S to XL, one line per
  model, for the prediction tasks, namely (\emph{a})~2D scalar, (\emph{b})~2D
  field, (\emph{c})~3D scalar, and (\emph{d})~3D field. Each panel is a
  cross-dataset aggregate.}
  \label{fig:trajectory_pred}
\end{figure*}

The benchmark's controlled scales from S to XL expose data-efficiency effects that
fixed-budget academic evaluations miss, as shown in Figure~\ref{fig:trajectory}.
In 3D geometry generation, DeepSDF is the most data-efficient generator
on the well-behaved DeepJEB and DeepWheel shapes, attaining the lowest FPD there
at the smaller scales. Its cross-dataset FPD in
Figure~\ref{fig:trajectory}(b) is nonetheless highest at scale~S, inflated by its
collapse on the non-watertight DrivAerNet cars, as detailed in
Section~\ref{sec:results:pertask}. BenchRank ranks it first at the three smaller
scales all the same, at~S even though PointFlow's aggregate FPD is far
lower at $18$ against $78$, because DeepSDF wins the rest of the metric
suite. PointFlow takes first at XL, where its broader distributional
coverage, with cross-dataset mean recall $0.40$ against $0.21$ and coverage $0.49$
against $0.35$, combines with a now-lower FPD. This is a case in which a single
headline metric and the full-suite ranking disagree. In 2D image generation, the
ordering inverts, since a VAE leads at S while the diffusion model DDPM
takes over from M onward, consistent with diffusion models' known appetite for
data. In 3D scalar regression, capacity-reduced variants lead almost throughout,
with Point Transformer-Small first at every scale but~M and PCT-Small
first on the DeepJEB structural response at~S, in a regime where their
full-capacity counterparts tend to collapse toward mean
prediction. The same scale-dependent orderings recur across the prediction tasks,
as shown in Figure~\ref{fig:trajectory_pred}. These crossovers are the kind of
finding the benchmark is built to surface.

The scaling curves also expose non-monotonic behavior that a converged
model should not exhibit. For several generators a quality metric worsens
as the training set grows. Examples are GAN (Basic) on the DeepJEB 2D images,
where FID rises from $164.3$ at M to $179.8$ at L, DeepSDF on DeepJEB 3D
shapes, where FPD rises monotonically from $7.5$ at S to $17.7$ at XL, and
3D-GAN, whose cross-dataset FPD rises from $38$ at L to $100$ at XL. Since
more data ought not to degrade a well-trained generator, we read these reversals
as symptoms of training instability rather than as data-efficiency findings. The
2D case survives repetition at further seeds, which makes it a property of the model
rather than of one draw. We therefore present
scale-dependent orderings as indicative, and flag the affected cells rather than
over-interpret them.

\subsection{Efficiency--Accuracy Trade-off}
\label{sec:results:efficiency}

\begin{figure*}[t]
  \centering
  \begin{minipage}{0.49\textwidth}\centering
    \includegraphics[width=\linewidth]{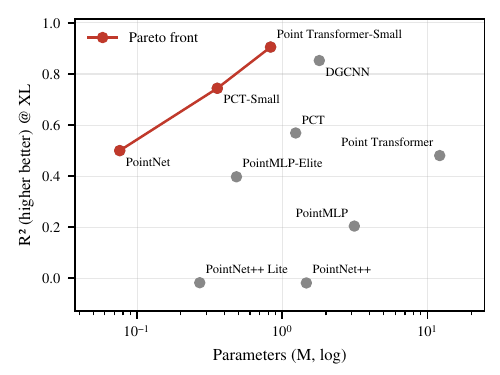}\\[-1pt]
    {\footnotesize(\emph{a}) 3D scalar prediction (DeepWheel)}
  \end{minipage}\hfill
  \begin{minipage}{0.49\textwidth}\centering
    \includegraphics[width=\linewidth]{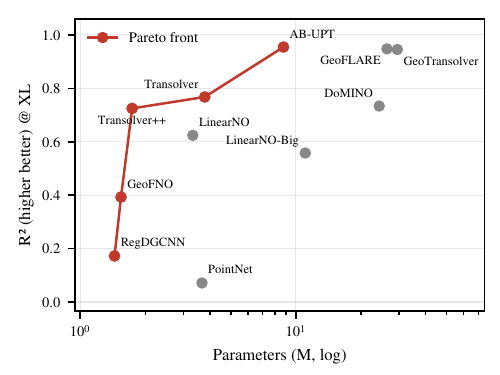}\\[-1pt]
    {\footnotesize(\emph{b}) 3D field prediction (DrivAerML)}
  \end{minipage}
  \caption[Parameters versus accuracy at scale~XL for two 3D tasks]%
  {Parameters versus accuracy, given as $R^2$ at scale~XL, for two 3D
  tasks. The red curve marks the Pareto frontier, with (\emph{a})~3D scalar
  prediction on \textbf{DeepWheel} and (\emph{b})~3D field prediction on
  \textbf{DrivAerML}.}
  \label{fig:pareto}
\end{figure*}

Because parameter count and run time are ranked metrics, the benchmark also
exposes the efficiency frontier in Figure~\ref{fig:pareto}, and the
efficiency view re-ranks models once cost is counted. The
frontier is task-dependent. In scalar regression, parameter count is a poor proxy
for utility in the small-data regime, and the capacity-reduced backbones
Point Transformer-Small at 0.8M and PCT-Small
at 0.4M sit on or near the frontier, dominating their full-capacity counterparts.
On DrivAerML 3D field prediction, where the large standalone operators compete on
a common dataset, the $\approx$26M GeoFLARE leads the ranking at every
scale while the larger $\approx$29M GeoTransolver follows and the
comparably large $\approx$24M DoMINO trails both, and the much smaller
official AB-UPT at $\approx$8.8M sits with the Transolver family well behind
them at scale~S before climbing past DoMINO into third place from M onward. At XL, AB-UPT
reaches the best pooled $R^2$ of the whole set at under a third of the parameters
of the models that outrank it, as detailed in Section~\ref{sec:results:pertask},
so larger is not reliably better even within one task. Cross-dataset efficiency must be read
with care, because models with partial dataset coverage are not directly
comparable, again as detailed in Section~\ref{sec:results:pertask}. The
cost-justified choice depends on the task and the data scale, which the efficiency
view makes explicit.

\subsection{Per-Task Results}
\label{sec:results:pertask}

We summarize each task. Full leaderboards for all datasets and scales, in both
the quality and the efficiency view, accompany the benchmark.

\begin{figure*}[tb]
  \centering
  \includegraphics[width=\textwidth]{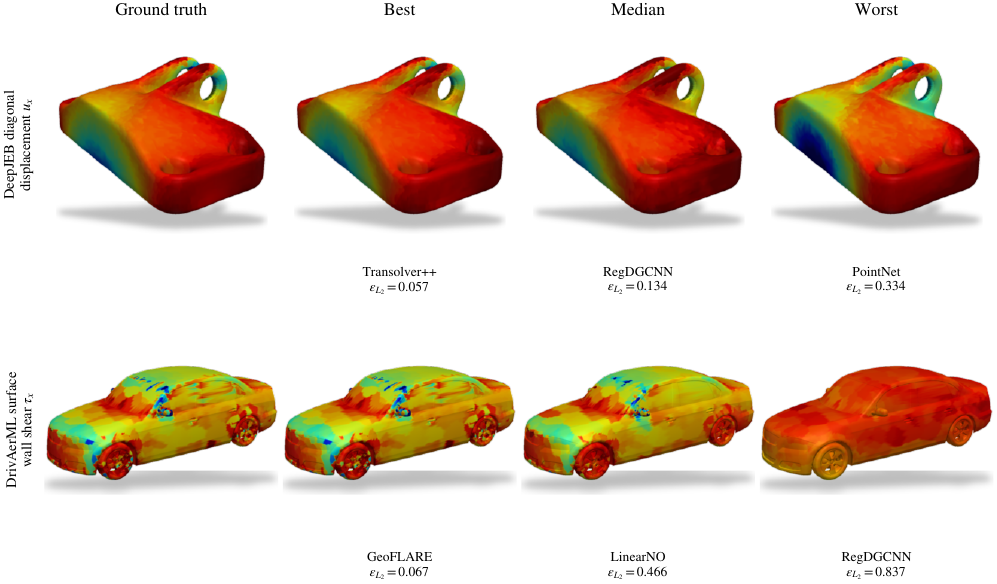}
  \caption[Qualitative 3D field prediction at scale~XL]%
  {Qualitative 3D field prediction at scale~XL. Each row shows the ground
  truth and the best-, median-, and worst-performing model on one held-out sample,
  for DeepJEB diagonal-load displacement $u_x$ on top and DrivAerML surface wall shear
  $\tau_x$ below. Models are ordered by their
  relative $L_2$ error over all components of the solve and labelled with
  $\epsilon_{L_2}$ for the component shown. Each row shares one jet color scale, and
  every model predicts the same seeded query set on the identical geometry.}
  \label{fig:qual_field3d}
\end{figure*}

\begin{figure*}[tb]
  \centering
  \includegraphics[width=\textwidth]{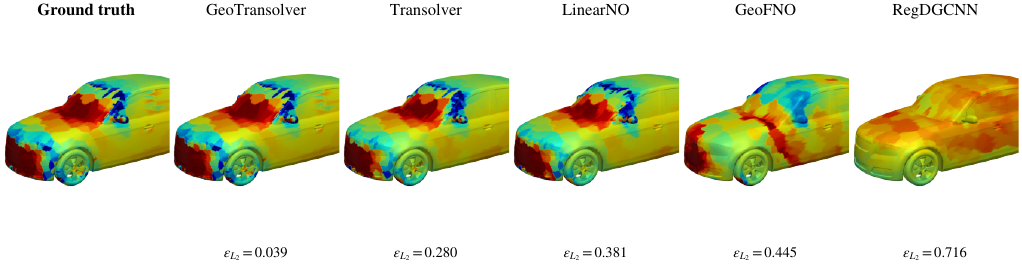}
  \caption[Local view of the DrivAerML surface-pressure case at scale~XL]%
  {Local view of the DrivAerML surface-pressure case at scale~XL. The
  ground truth is at left, followed by five models ordered best to worst from left
  to right by relative $L_2$ error.}
  \label{fig:qual_zoom3d}
\end{figure*}

\begin{figure*}[tb]
  \centering
  \includegraphics[width=\textwidth]{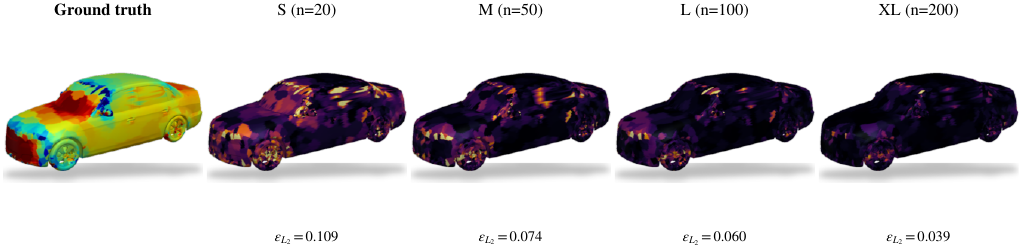}
  \caption[Field prediction error versus data scale on DrivAerML surface pressure]%
  {Field prediction error versus data scale on the DrivAerML
  surface-pressure case. The ground-truth pressure field is at left in jet, and the
  remaining panels map the absolute error $|\mathrm{pred}-\mathrm{GT}|$ of
  GeoTransolver on the same held-out car for the model trained at scale~S through
  XL, on one shared error scale where darker is lower error. Each panel is labelled
  with its relative $L_2$ error $\epsilon_{L_2}$.}
  \label{fig:progression3d}
\end{figure*}

\subsubsection{Field Prediction}
\begin{figure*}[tb]
  \centering
  \includegraphics[width=\textwidth]{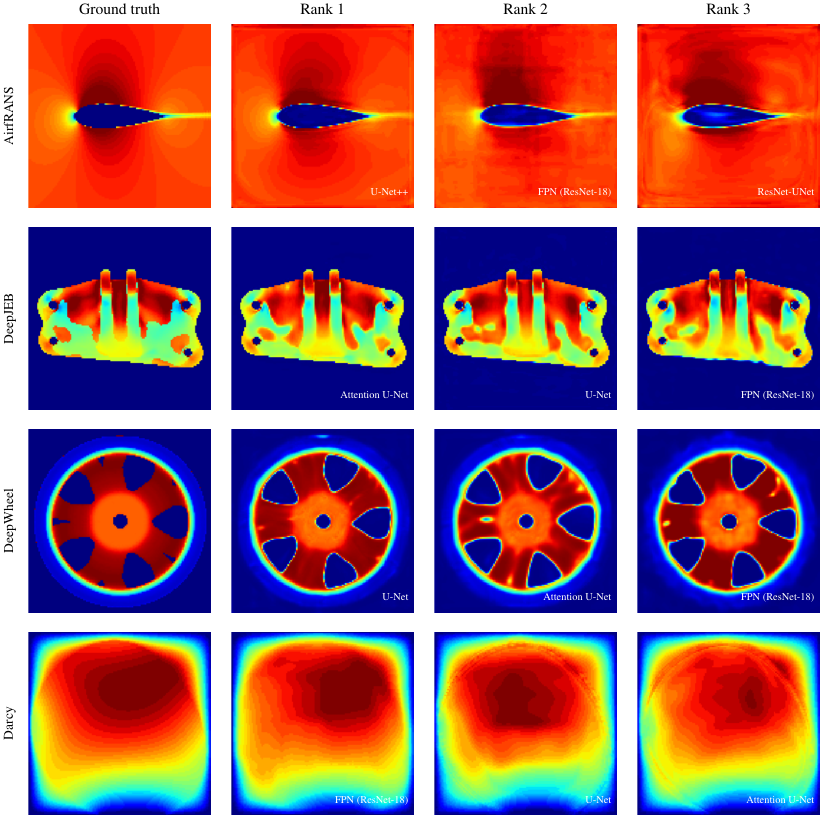}
  \caption[Qualitative 2D field prediction at scale~XL]%
  {Qualitative 2D field prediction at scale~XL. Each row shows ground truth
  and the top-three ranked models on one test sample per dataset, for AirfRANS:
  streamwise velocity $u$, DeepJEB: stress, DeepWheel: depth, and Darcy: pressure.
  Ranks are those of the field shown, not of the dataset aggregate over all of its
  targets. Each row shares a jet color scale, and the model is named in each panel.}
  \label{fig:qual_field2d}
\end{figure*}

3D field prediction must be read per dataset, because four of its models have
non-uniform coverage. The standalone GeoFLARE, GeoTransolver, DoMINO, and the
official AB-UPT are evaluated on DrivAerML only, which is
the automotive-aero dataset they were developed for, while the Transolver family,
RegDGCNN, GeoFNO, LinearNO, and PointNet are evaluated across all 3D-field
datasets. On DrivAerML, where these models compete on a common dataset,
GeoFLARE leads at every scale, as shown at scale~S in Table~\ref{tab:lb:field3d},
with GeoTransolver second throughout. DoMINO is third at that scale and AB-UPT
takes the place from M onward, with the Transolver family behind both. The
leading three separate decisively, whereas ranks six through eight sit within
$0.75\%$ and $0.36\%$ of total score of one another, and the three models below
the printed top-eight cut fall within $1.6$ points of rank six. That band is a
near-tie of the kind \ref{app:robustness} shows a reweighting can reorder, so we
do not read an order within it. GeoFLARE and GeoTransolver share a geometry
encoder and differ in how the latent tokens mix, since GeoFLARE substitutes FLARE
low-rank self-attention for the physics-attention slices of its sibling. The
substitution wins on accuracy at $26.41$M parameters against $29.49$M and at
broadly comparable training cost, roughly two thirds of its sibling's on the
surface configuration but about a quarter more on the volume one, which makes it
the clearest case in the benchmark of an attention mechanism, rather than
capacity or geometry handling, carrying the improvement. That advantage is
largest where data is scarcest. Averaged over the DrivAerML cells, GeoFLARE's
improvement in relative $L_2$ over its sibling falls monotonically from $8.4\%$ at
scale~S to $6.4\%$ at M, $4.4\%$ at L, and $2.2\%$ at XL, so the newer attention
buys most in the small-data regime and the two converge as data accumulates. This DrivAerML board pools the surface and volume
configurations, so its MAE and RMSE average components of differing physical units
and are read as relative magnitudes rather than a single physical error. The pooled
ordering also hides a split between the two configurations. On volume,
GeoFLARE is first at every scale. On surface it is first at the two
smaller scales, whereas GeoTransolver recovers the lead at L and XL even
though GeoFLARE retains the better squared error there, the two being
separated by at most $0.001$ in $R^2$ and the ranking turning instead on percentage
error and sign agreement. At the largest scale AB-UPT attains the best
pooled $R^2$ of $0.955$, against $0.946$ for GeoTransolver and $0.948$ for
GeoFLARE, yet places third once the full metric set is aggregated, a
reminder that a single accuracy statistic and a multi-metric ranking need not
agree. All models are scored on the same seeded $8192$-point query set, but their
geometry encoders ingest their published input. The Transolver family,
GeoFNO, LinearNO, PointNet, and AB-UPT encode the same $8192$-point sample,
whereas GeoFLARE and GeoTransolver sample $50$k surface points and
DoMINO encodes the full mesh, so the common accuracy axis compares
predictions at identical query points, not identical geometry-input budgets.
AB-UPT's largest-scale accuracy is therefore reached at the same
$8192$-point budget as the transformers ranked above it and is not an
input-resolution artifact, while GeoFLARE, GeoTransolver, and
DoMINO additionally benefit from the richer geometry their architectures
are built to consume. On the DeepJEB structural load cases and DrivAerNet center-plane, the
Transolver family and RegDGCNN lead the common-coverage set. We therefore do
not read the pooled cross-dataset aggregate, in which GeoFLARE,
GeoTransolver, and AB-UPT top the high-data scales, as evidence of
uniform superiority. Their pooled rank reflects narrower, easier coverage that
omits the hardest DeepJEB cases. Across the 3D-field datasets, the gap between operators and
point/graph baselines is largest on the coupled displacement/stress and modal
problems, where the Modal Assurance Criterion and sign/extremal-agreement metrics
separate physically faithful predictions from those that merely minimize pointwise
error. The two co-produced quantities are also not equally learnable. Every model
attains a lower relative error on displacement than on stress in all sixteen
combinations of four load cases and four data scales, and at scale~XL the best model
reaches $R^2$ of $0.93$ to $0.98$ on displacement against only $0.76$ to $0.83$ on
stress, so a target-level average of the two reports an accuracy neither channel
attains, which the component ranking level exists to expose.
On the modal problem every model scores $R^2<0$, worse than mean
prediction, at all four scales, and the task is never learned regardless of data
size. We deliberately report and rank these cells rather than hide them. An
all-$R^2{<}0$ result is itself decision-relevant, since it tells a practitioner
that no model in the pool solves this task at this scale, while the
relative order among the failed models is flagged as indicative only.
Figure~\ref{fig:qual_field3d} contrasts the best-, median-, and worst-ranked model
on one DeepJEB displacement and one DrivAerML wall-shear case. The best model
tracks the reference almost exactly on the raw engineering geometry, at relative
$L_2$ error ${\approx}0.06$ and $0.07$ for the two rows, while the worst washes out
the field structure entirely, so the leaderboard ordering is visible directly and
not only in the aggregate metrics. In the wall-shear row that failure is unmistakable,
since the worst model returns a near-uniform field and loses
both the high-shear bands over the windscreen and roof and the reversed-flow patches
at the wheel wakes and the base, which the sign of $\tau_x$ separates. The two rows are
the diagonal load case, which is the hardest of the DeepJEB cases for the best model
at every scale, and wall shear, which is harder than the pressure field of the same
solve for every model in the pool, so the figure understates rather than flatters
achievable accuracy. Figure~\ref{fig:qual_zoom3d} zooms the DrivAerML
surface-pressure case to the front region and shows that the residual is structured,
concentrated at the front-stagnation region and wheels and recovered only by the
strongest models. The analogous DeepJEB residual concentrates at the bracket bolt
holes. Figure~\ref{fig:progression3d} isolates the data-scale axis on the same
DrivAerML surface-pressure case, where one of the two strongest models,
GeoTransolver, already
predicts the field well at every scale, so the gain shows most clearly in the
absolute-error map, which contracts as data grows while the relative $L_2$ error
falls roughly threefold, from $0.11$ at scale~S to $0.04$ at XL, the qualitative
counterpart of the $R^2$-versus-scale trajectory in
Figure~\ref{fig:trajectory_pred}(d). In 2D field prediction, five of the eight models
take the top spot at some scale or dataset, with U-Net++, Attention U-Net, and FPN
(ResNet-18) trading it most often. Achieved accuracy varies widely across the four datasets even
at the smallest scale. At~S the best model reaches $R^2{=}0.96$ on the
single-channel AirfRANS flow field $p/\rho$ and $0.93$ on the structural stress
field, $0.82$ on Darcy pressure, and only $0.66$ on
monocular depth for DeepWheel. Because each target is normalized differently,
and because
depth is scored on a foreground mask as described below, these $R^2$ are an
achieved-accuracy spread, not a strict cross-dataset difficulty ranking. For
depth, the regression metrics are computed on a foreground mask so that the large
constant background cannot inflate scores. The background is roughly half the
frame, so without this correction a model is rewarded for trivially predicting it,
which illustrates how an engineering-appropriate metric definition changes the
ranking. The DeepJEB structural problem additionally exposes a displacement target
alongside stress. The same U-Net and FPN leaders learn it cleanly, with six of the
eight models reaching $R^2{>}0.9$ at scale~XL, while DPT-Hybrid trails
the field at $0.37$ after scoring below zero on displacement at every
smaller scale, down to $R^2{\approx}-8$ at~S. The
2D-field task also yields one of the benchmark's sharpest reversals of
conventional wisdom, with a nuance a single-task evaluation would miss. The
smallest model is the most reliable, and the largest pretrained ones are erratic.
The 7.8M from-scratch U-Net is uniformly strong across all four
2D-field datasets, reaching $R^2{=}0.57$--$0.99$, whereas the 122.4M
depth-pretrained DPT-Hybrid, a MiDaS model, and the 61.2M GLPN
are erratic across channels. Both collapse to $R^2{<}0$ at every scale on the
Darcy-pressure field, down to $R^2{\approx}-10$, and on the AirfRANS
eddy-viscosity channel, far worse than predicting the mean field, and DPT-Hybrid
additionally falls below zero on both DeepJEB structural channels at the smaller
scales. Each is nonetheless strong on some targets, GLPN reaching
$R^2{>}0.9$ on the AirfRANS velocity components at the larger scales and
DPT-Hybrid taking the top rank on monocular depth at~S, so neither is uniformly
weak and neither is uniformly reliable. In this small-data engineering-field
regime, parameter count and large-scale natural-image pretraining are not
generic advantages, since they help only within-domain. This is a direct inversion
of the dense-prediction intuition that bigger, depth-pretrained transformers
transfer best, and exactly the kind of result a single-task academic evaluation
would not surface. Figure~\ref{fig:qual_field2d} shows ground-truth and predicted
fields for the four 2D-field datasets, on the real geometries the benchmark
targets, showing for AirfRANS the streamwise velocity, the most legible of its four
channels, and for DeepJEB the stress field, the harder of its two structural
targets.

\begin{table*}[t]
  \centering\footnotesize
  \caption[3D field prediction on DrivAerML at scale~S]%
  {3D field prediction on \textbf{DrivAerML} at scale~S, quality view,
  top~8, pooled over the surface and volume configurations. AB-UPT is reported
  under a research license and excluded from the public leaderboard.}
  \label{tab:lb:field3d}
  \resizebox{\textwidth}{!}{\begin{tabular}{@{}c l r r r r r r r r r@{}}
\toprule
Rank & Model & Total Score & MAE $\downarrow$ & RMSE $\downarrow$ & MAPE (\%) $\downarrow$ & $R^2$ $\uparrow$ & Rel-L2 $\downarrow$ & MAC $\uparrow$ & Sign Agree $\uparrow$ & Extremal Agree $\uparrow$ \\
\midrule
1 & GeoFLARE & 100.00 & 2.78 & 4.88 & 48.02 & 0.88 & 0.28 & 0.97 & 0.89 & 0.87 \\
2 & GeoTransolver & 89.03 & 3.15 & 5.58 & 50.04 & 0.87 & 0.30 & 0.97 & 0.89 & 0.86 \\
3 & DoMINO & 57.37 & 10.51 & 18.29 & 77.53 & 0.62 & 0.54 & 0.71 & 0.81 & 0.67 \\
4 & AB-UPT & 40.29 & 12.62 & 18.77 & 116.06 & 0.41 & 0.72 & 0.63 & 0.71 & 0.47 \\
5 & Transolver++ & 39.00 & 13.86 & 21.28 & 116.54 & 0.36 & 0.74 & 0.60 & 0.70 & 0.42 \\
6 & LinearNO-Big & 36.21 & 12.99 & 20.13 & 138.30 & 0.23 & 0.82 & 0.54 & 0.64 & 0.33 \\
7 & Transolver & 35.94 & 13.32 & 20.48 & 136.87 & 0.24 & 0.82 & 0.54 & 0.64 & 0.34 \\
8 & LinearNO & 35.81 & 14.54 & 21.87 & 137.76 & 0.23 & 0.82 & 0.54 & 0.64 & 0.37 \\
\bottomrule
\end{tabular}
}
\end{table*}

\subsubsection{Scalar Prediction}
\label{sec:results:scalar}
\begin{figure}[tb]
  \centering
  \includegraphics[width=\colfigw]{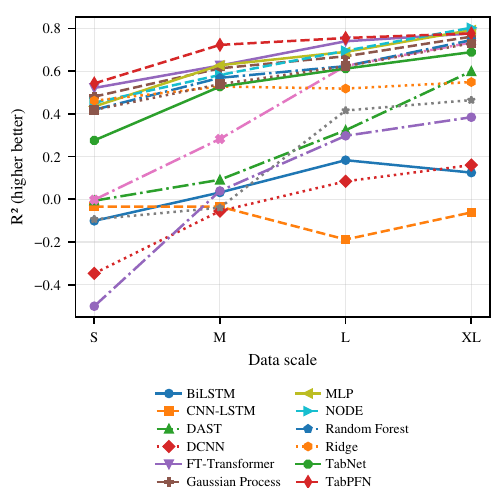}
  \caption[Primary accuracy versus data scale for 1D scalar prediction]%
  {Primary accuracy $R^2$ versus data scale for 1D scalar prediction, one
  line per model. The tabular models are scored on Concrete and Airfoil pooled and
  the time-series models on CMAPSS, so the two pools share the axis without
  competing.}
  \label{fig:trajectory_scalar1d}
\end{figure}

Scalar regression most sharply exposes small-data behavior. In 3D, on DeepWheel
mass and modal, a compact point backbone leads at scale~S, namely a
capacity-reduced Point-Transformer variant on both, at $R^2{\approx}0.60$,
while full-capacity transformers collapse toward mean prediction. By scale~XL the
same tasks are largely solved, at $R^2{\approx}0.93$ for mass and $0.89$ for
modal, a data-efficiency trend overall, though the intermediate scales remain the
least settled stretch of the curve in Figure~\ref{fig:trajectory_pred}(d). The same
capacity-driven pattern recurs on the \textbf{DeepJEB} coupled structural
response, where displacement and stress are predicted jointly from a single
checkpoint, from 2D-image and 3D-point-cloud inputs. Compact point and image
backbones lead, reaching $R^2{\approx}0.9$ in 3D at scale~XL, while the
full-capacity PointNet++ collapses to mean prediction at the three smaller scales
and recovers only to $R^2{=}0.24$ at XL. The 1D tabular tasks show a
complementary pattern. On Concrete at $n{=}20$, the in-context foundation model
TabPFN leads at $R^2{=}0.61$, with FT-Transformer, random-forest, and
Gaussian-process close behind, whereas LightGBM collapses to the mean at
$R^2{\approx}0$ and falls below the top-8 shown in Table~\ref{tab:lb:concrete}. The
other gradient-boosted baseline, XGBoost, reaches $R^2{=}0.51$, and the
from-scratch deep models MLP and TabNet trail at $R^2{=}0.35$ and $0.31$, so the
small-sample regime inverts the usual tabular ordering. On Airfoil the lead rotates
through Ridge, TabPFN, FT-Transformer, and the MLP as the scale grows.
Across the pooled tabular scales the top model shifts from TabPFN at the three
smaller sizes to the deep NODE model at XL, as the per-scale trajectories in
Figure~\ref{fig:trajectory_scalar1d} show, and no gradient-boosted model leads at
any scale, because even the XL size of 200 rows stays below the regime where
boosted trees overtake in-context and deep learners.
On CMAPSS remaining-useful-life, no model exceeds $R^2{\approx}0$ at scale~S. The
best sit near zero and the weakest fall to $R^2{\approx}-0.5$, because 20
run-to-failure engines is barely above mean prediction for a $543{\times}24$
time-series, with separation emerging only at larger scales. Here it is the
rank-correlation metrics Pearson and Spearman, which a constant predictor cannot
satisfy, that prevent a mean-collapsing model from appearing competitive on
error-only metrics. The viability gate itself is configured for the generation
tasks, as described in Section~\ref{sec:method:benchrank}, and does not act on
these regression cells.

\begin{table*}[t]
  \centering\footnotesize
  \caption[1D tabular regression on Concrete at scale~S]%
  {1D tabular regression on Concrete at scale~S, with $n{=}20$, quality
  view, top~8, showing the complete ranked metric set for scalar prediction.}
  \label{tab:lb:concrete}
  \resizebox{\textwidth}{!}{\begin{tabular}{@{}c l r r r r r r r r r@{}}
\toprule
Rank & Model & Total Score & MAE $\downarrow$ & RMSE $\downarrow$ & MAPE (\%) $\downarrow$ & $R^2$ $\uparrow$ & Rel-L2 $\downarrow$ & MaxAE $\downarrow$ & Pearson $\uparrow$ & Spearman $\uparrow$ \\
\midrule
1 & TabPFN & 100.00 & 8.43 & 10.38 & 27.07 & 0.61 & 0.27 & 31.83 & 0.79 & 0.79 \\
2 & FT-Transformer & 55.30 & 8.64 & 10.91 & 27.44 & 0.57 & 0.28 & 34.63 & 0.78 & 0.80 \\
3 & Random Forest & 44.53 & 9.19 & 11.32 & 29.69 & 0.54 & 0.29 & 33.69 & 0.76 & 0.75 \\
4 & Gaussian Process & 44.19 & 9.09 & 11.53 & 29.23 & 0.52 & 0.29 & 33.17 & 0.73 & 0.73 \\
5 & NODE & 41.19 & 9.27 & 11.61 & 29.56 & 0.52 & 0.30 & 37.91 & 0.72 & 0.74 \\
6 & XGBoost & 41.10 & 9.20 & 11.74 & 29.80 & 0.51 & 0.30 & 39.28 & 0.75 & 0.77 \\
7 & Ridge & 39.59 & 9.54 & 12.29 & 29.84 & 0.46 & 0.31 & 44.29 & 0.69 & 0.73 \\
8 & MLP & 37.52 & 10.62 & 13.50 & 34.77 & 0.35 & 0.35 & 41.85 & 0.64 & 0.64 \\
\bottomrule
\end{tabular}
}
\end{table*}

\subsubsection{Generation}
\begin{figure}[tb]
  \centering
  \includegraphics[width=\colfigw]{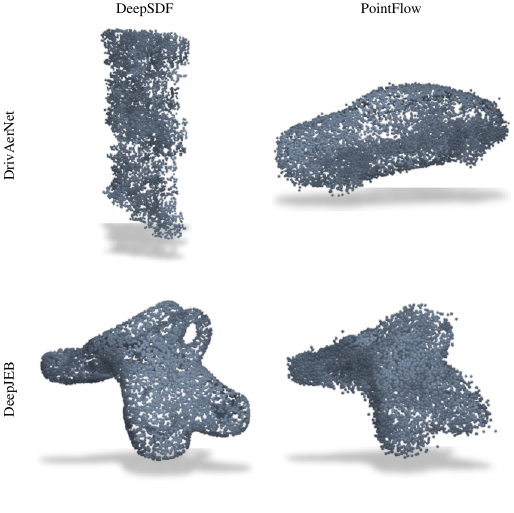}
  \caption[Qualitative 3D geometry generation at scale~S]%
  {Qualitative 3D geometry generation at the smallest scale, with
  $n{=}20$. Point clouds from the SDF auto-decoder DeepSDF and the
  continuous-normalizing-flow generator PointFlow on two datasets, namely
  DrivAerNet cars and DeepJEB brackets.}
  \label{fig:qual_gen3d}
\end{figure}

The generation tasks display the clearest data-scale crossovers. In 2D image
generation, a VAE ranks first at scale~S on all three datasets, but the diffusion
model DDPM takes the top rank from M onward and at XL. This is a reminder
that BenchRank rewards distributional coverage and diversity, not FID alone, so
the best single-metric model is not always the best-ranked one. DeepWheel is the
hardest 2D domain throughout. Its thin spoke structures, chrome specularity, and
near-achromatic renders keep FID high for every model, and the top-ranked model's
FID stays $\approx$143 at XL against $\approx$79 on DrivAerNet, which the benchmark
surfaces as an intrinsic data difficulty rather than a model defect. In 3D
geometry generation, the SDF model DeepSDF is the most data-efficient on
the well-behaved DeepJEB and DeepWheel shapes, with the lowest FPD at the smaller
scales, while PointFlow tops the cross-dataset ranking at XL. The most
striking case is DrivAerNet. Its non-watertight car meshes give the
signed-distance field an ill-posed inside/outside target, and because
DeepSDF generates by sampling a Gaussian mixture fit to its learned
latent codes, as described in Section~\ref{sec:method:pipeline}, at scale~S that
mixture is estimated from only twenty shapes. The two effects together collapse
its samples to a flat sheet at FPD $\approx$221, whereas PointFlow, which
has a native sampling prior, stays robust at FPD $\approx$9. This is a data-starved
generation failure, not a failure to reconstruct, and the per-scale,
per-dataset leaderboard surfaces it directly. Figure~\ref{fig:qual_gen3d} makes
the contrast visible. On DrivAerNet DeepSDF collapses to a flat sheet
while PointFlow recovers a car, yet on the well-behaved DeepJEB bracket
DeepSDF is the cleanest generator. Beyond FPD, the two structural metrics
expose independent failures. On DeepWheel AtlasNet keeps Manifold-$\Delta$
low at $\approx\!0.01$--$0.02$, giving locally clean $2$-manifold patches, yet its
Uniformity-$\Delta$ runs over an order of magnitude higher than DeepSDF's,
at $0.23$--$0.30$ against $\le\!0.01$, because its patch decoder clumps points
along seams. This is a non-uniformity that FPD and Manifold-$\Delta$ both miss and
that Uniformity-$\Delta$ is designed to expose.

\subsection{Academic vs.\ Industrial Performance}
\label{sec:results:gap}
\begin{figure*}[tb]
  \centering
  \includegraphics[width=\textwidth]{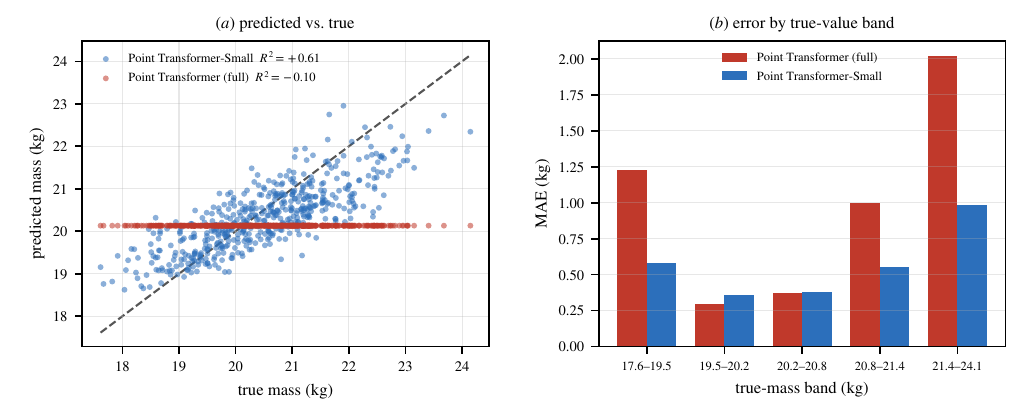}
  \caption[Capacity-driven mean collapse on DeepWheel mass at scale~S]%
  {Capacity-driven mean collapse on DeepWheel mass at scale~S, for
  Point Transformer at 12.1M parameters against its capacity-reduced sibling
  Point Transformer-Small at 0.8M.}
  \label{fig:collapse}
\end{figure*}

Taken together, the results show that an architecture's standing at large academic
scale is a weak predictor of its standing in the small-data, real-geometry regime
that engineers face. We take that academic standing as reported in the originating
publications, which we treat as a qualitative reference rather than re-measure
here. Architectures that headline academic leaderboards are frequently not
the best choice at scale~S. These include diffusion models for generation,
high-capacity or heavily pretrained transformers for regression, and implicit SDF
generators on watertight benchmark shapes. At scale~S they are beaten by a VAE, by
capacity-reduced, classical, or from-scratch convolutional models, and by a flow
model on non-watertight geometry, respectively. Figure~\ref{fig:collapse}
illustrates the mechanism on one controlled pair. The full-capacity
Point Transformer and its capacity-reduced sibling
Point Transformer-Small differ only in width, yet on DeepWheel mass at scale~S the
full model collapses to the mean and ignores the geometry, emitting a prediction
with no variance at all, while the smaller model recovers the true value, with the
error concentrated in the tails where a mean prediction is most wrong. We attribute this primarily to the
data-scale effect, which we vary directly. We do not claim to isolate a
distinct ``industrial-domain'' effect, since scale and data domain are not varied
independently here, and we run no academic-scale or academic-domain control on the
same models. Disentangling them is left to future work. The actionable conclusion
is conservative but firm. Rankings do not transfer across data scale and must be
measured. This is the core of \benchname{}'s decision-support thesis. Rather than a
universal champion, the benchmark delivers a per-task, per-scale recommendation
grounded in the conditions engineers actually face, and a reproducible procedure
for re-deriving it as data, models, and scales change.

\subsubsection*{Anti-Trend Cases Are Genuine} A few models buck the prevailing
trend in which more data yields better results, and we audited each to confirm it
is genuine model behavior in this regime rather than a pipeline artifact. On 2D
fields, the depth-pretrained GLPN collapses to negative $R^2$ on the
Darcy-pressure field and on the AirfRANS eddy-viscosity channel at every scale,
even while it reaches $R^2{>}0.9$ on the AirfRANS velocity components at the
larger scales and stays mid-pack on the DeepJEB structural channels. This is genuine
per-field negative transfer on those targets, not a loading fault, and its
pretrained weights load correctly. On the smallest 3D scalar tasks the
full-capacity Point Transformer and PCT transformers collapse toward mean
prediction, and the reduced Point Transformer variant dips at the intermediate
scale~M before recovering. This is the capacity-data mismatch that motivates the
capacity-reduced variants we include. In generation, 3D-GAN degrades at the largest 3D
scale owing to GAN training instability, and weight-clipping WGAN-CP
trails at the larger 2D scales, the documented weakness of weight clipping
relative to gradient-penalty WGANs. In every case the per-scale leaderboard
surfaces the behavior rather than hiding it, as seen in
Figure~\ref{fig:trajectory_pred}.

\subsubsection*{Worked Example}
Concretely, a practitioner facing the DeepJEB structural field problem with
roughly 100 high-fidelity simulations, which is the L bucket, reads that dataset's
own 2D-field L leaderboard. On this dataset, and unlike the pooled winner reported
in Table~\ref{tab:winners}, FPN (ResNet-18) with $15.6$M parameters ranks
first in the quality view, ahead of the $8.0$M Attention U-Net and the $7.8$M
U-Net, so it is the default recommendation. Notably the FPN's lead comes from
higher structural similarity, with SSIM $0.89$ against $0.84$ and $0.82$, even
though Attention U-Net has marginally better pixel PSNR at $28.8$ against $27.9$,
a marginally better $R^2$ at $0.95$ against $0.94$, and equal MAE $0.02$. This is
exactly the field-fidelity-over-pixel-error tradeoff the suite is built to expose.
If that model must run inside a design-optimization loop on a single workstation
GPU, the practitioner switches to the efficiency view, which counts
parameters and run time. That view reorders the same three models, promoting
Attention U-Net to first and U-Net to second while FPN falls to third, because
the two smaller models match or beat FPN on every ranked quality metric except
SSIM, at half the parameters. The
benchmark does not return ``the best model'' in the abstract. It
returns the model whose accuracy and cost fit this task at this scale, and it lets
the practitioner re-run the recommendation as the dataset grows. We stress that the chosen margin, meaning how much accuracy a
practitioner will trade for efficiency, is a deployment decision the leaderboard
informs but does not make.

\section{Conclusion}
\label{sec:conclusion}

We introduced \benchname{}, a unified benchmark and leaderboard, deployed
as the \sitebrand{}, that evaluates generative and predictive models under a
single standardized procedure on industrial engineering datasets, across seven
tasks spanning 1D, 2D, and 3D domains and a controlled range of data scales.
\benchname{} couples industrial and public CAD/CFD/FEA data with a common metric
suite and BenchRank, a debiased graph-based ranking in which every reported
quality metric is also a ranked metric and computational cost is reported
separately in an efficiency view. Each family is ranked within its own tasks
rather than head-to-head against the other. The central findings are that no
single model dominates across tasks and that the best model frequently changes
with the available data scale. A useful recommendation is therefore conditioned
on both the task and the data budget rather than on a universal best model. By
measuring competing models under common industrial conditions with a debiased
and reproducible ranking, \benchname{} turns ``state-of-the-art'' from a
self-reported claim into an openly published result that engineers
can build on.

\subsection{Limitations}
Several limitations bound the present study and frame its claims.
\textbf{Statistical.} The reported value is the mean over a cell's repeated training
runs, and that basis is fixed per task rather than per
model, since ranking averaged models against un-averaged ones would remove the
run-to-run luck from some competitors and not others. Five of the seven tasks are
reported on that mean, and 3D generation and 3D field prediction fall back to the
single baseline run because their rerun sets are incomplete, as
Section~\ref{sec:method:pipeline} details. The leaderboards carry no
per-cell confidence intervals, and the training run is examined separately in
\ref{app:robustness}. Closely-spaced ranks may therefore lie within run-to-run
noise, and those crossover narratives should be read as indicative. Several generators also show
non-monotonic scaling, in which a quality metric worsens as the training
set grows, as reported in Section~\ref{sec:results:scaling}. A converged model
should not exhibit this behavior. Where the reruns exist, the effect survives
averaging and is therefore a property of the model rather than of one draw. Where
they do not, part of the swing may still be run-to-run. We deliberately
retain these anomalies in the published results rather than smoothing them over,
and broadening the repeated-run coverage remains open work.
\textbf{Ranking.} BenchRank's hyperparameters, namely target dominance-graph
density, redundancy strength, PageRank damping, and the viability-gate
threshold, are fixed by construction. A post-hoc sensitivity study in
\ref{app:robustness} shows that the rankings are largely insensitive to
them, with mean Kendall $\tau{=}0.98$ under $\pm$perturbation and rank-1 stable
in $95\%$ of cells, and positively but not perfectly correlated with simpler
normalized-mean and Borda aggregators. That study is post-hoc and does not cover
an Elo-style aggregator or the viability-gate threshold. The
$[0,100]$ Total Score is an ordinal, within-task-and-scale normalization
anchored at 100 for the top model, not a cardinal or cross-task-comparable
quality measure.
\textbf{Coverage.} Some tasks rest on a single dataset, a few coupled problems are
scored on a single output channel, and the smallest scales can leave certain
tasks near mean prediction, as in remaining-useful-life regression with only
tens of trajectories. Four standalone 3D-field models, GeoFLARE, GeoTransolver,
DoMINO, and the official AB-UPT, are evaluated on DrivAerML only, so the pooled
cross-dataset 3D-field aggregate mixes models of differing coverage and must be
read per dataset, as detailed in Section~\ref{sec:results:pertask}. License
constraints exclude some recent models from the public leaderboard,
though the models remain fully evaluable for research and we report their
measured metrics. A substantial fraction of the evaluation cells, all of the
DeepWheel and DrivAerNet tasks, rests on CC~BY-NC data. This license permits the
academic evaluation reported here but not commercial redistribution, so while
the \benchname{} methodology and evaluation procedure transfer directly
to commercial practice, a commercial adopter would re-apply them to its own or
permissively-licensed data before deploying the resulting models.
\textbf{Governance.} As a vendor-hosted ``living'' leaderboard whose authors also
contribute datasets and, internally, models, the platform depends for its
long-term value on maintenance, versioned ``as-of'' rankings, and anti-gaming
validation of contributed results. We exclude all author-affiliated in-house
models from publication regardless of their measured rank, so the policy applies
whether an in-house model would have won or lost, but a fuller governance and
conflict-of-interest policy is needed.
Finally, rankings reflect the fixed training budgets and metric suite adopted
here, and alternative budgets or metrics could reorder closely-spaced models.

\subsection{Future Work}
Natural extensions include broadening dataset and task coverage, multi-channel
field prediction, surface-cleanliness and downstream CAD/B-rep convertibility
metrics for generated geometry, continuously updated rankings as new models are
contributed, and moving from passive ranking toward an agentic recommendation
that selects a model given a task, a dataset, and resource constraints. Among
these, the agentic recommendation is the most immediately promising direction for
future research, since it turns a published ranking into an answer to the question
an engineer actually asks. Widening the repeated-run coverage would place every
task's scale-dependent findings on the same statistical footing.

\section*{Acknowledgment}
This work was supported by grants from the Ministry of Science and ICT (GTL24033-000, N10250154, and No.~2022-0-00986), the Ministry of Trade, Industry and Energy (RS-2025-02317327 and RS-2025-25444634), the Ministry of Oceans and Fisheries (PET0050), and Korea Hydro \& Nuclear Power Co., Ltd. (No.~8-Tech-07).

\section*{Conflict of Interest}
There are no conflicts of interest.

\section*{Data Availability}
The public leaderboard is hosted at \url{https://leaderboard.narnia.ai}. The data behind it, namely the per-model benchmark scores and per-metric values for every ranked run, are publicly available at \url{https://github.com/Narnialabs/leaderboard}. Those files label each model by the short identifier under which its results are released, and the map from those identifiers to the display names used here ships with the released data. The one exception is AB-UPT, whose research license withholds it from the release, so the 3D field-prediction boards in this paper rank a model that the released files do not contain.

\appendix
\renewcommand{\thesection}{Appendix~\Alph{section}}
\section{Metric Definitions}
\label{app:metrics}

This appendix gives precise definitions of the quality metrics that
Section~\ref{sec:method:metrics} references but does not spell out. These are both
the engineering-specific validity metrics \benchname{} introduces or adapts, whose
motivation and necessity are argued in Section~\ref{sec:method:metrics}, and the
less-familiar standard metrics adopted from the generative-modeling and vision
literature. Throughout, let $N$ be the number of points or query locations in a
sample, let $\hat{y}$ be the prediction, and let $y$ be the ground truth. Norms
and dot products are taken per sample, and results are averaged over the sample
set. The first five entries are the engineering-specific validity metrics, namely
two pattern-quality measures of sign and extremal agreement, two
structural-quality measures of manifold- and uniformity-$\Delta$, and the vector
and modal MAC. The remainder are the standard adopted metrics.

\begin{description}
\item[Sign Agreement ($\uparrow$).] The fraction of points at which the predicted
  and true fields share sign, written
  $\frac{1}{N}\sum_{i} [\,\mathrm{sign}(\hat{y}_i) = \mathrm{sign}(y_i)\,]$ with an
  Iverson bracket, which measures whether the field loads in the correct
  direction.
\item[Extremal Agreement ($\uparrow$).] Recall of the peak region. With
  $k = \lceil 0.1N \rceil$ and $T_k(\cdot)$ the index set of the $k$
  largest-magnitude entries, it is $|T_k(\hat{y}) \cap T_k(y)| / k$, which measures
  whether the predicted stress and pressure concentrations coincide with the true
  top-$10\%$ hot spots.
\item[Manifold-$\Delta$ ($\downarrow$).]
  $|t_{\mathrm{gen}} - t_{\mathrm{real}}|$, where the local ``thinness''
  $t = \mathrm{mean}_i\, \lambda_3^{(i)} / (\lambda_1^{(i)} + \lambda_2^{(i)} +
  \lambda_3^{(i)})$ averages the smallest normalized PCA eigenvalue over each
  point's $k$-NN neighborhood with $k{=}20$. A clean $2$-manifold surface gives
  $t \to 0$ and an isotropic blob gives $t \to 1/3$, so the $\Delta$ penalizes
  generated shapes whose local dimensionality departs from the real data.
\item[Uniformity-$\Delta$ ($\downarrow$).]
  $|c_{\mathrm{gen}} - c_{\mathrm{real}}|$, where $c$ is the coefficient of
  variation of $k$-NN distances with $k{=}8$, a proxy for how evenly points are
  sampled over the shape. It is complementary to Manifold-$\Delta$. A generator
  can place points on a locally clean surface, giving low Manifold-$\Delta$, yet
  clump them along seams, giving high Uniformity-$\Delta$, which leaves
  under-resolved regions that hinder downstream meshing and simulation, so the two
  are reported separately.
\item[Modal Assurance Criterion (MAC, $\uparrow$).] For a vector or modal field
  with predicted and reference vectors $\hat{\phi}$ and $\phi$, this is the squared
  cosine similarity
  $\mathrm{MAC} = |\hat{\phi}\cdot\phi|^{2} / (\|\hat{\phi}\|^{2}\,\|\phi\|^{2})$.
  It is invariant to sign and scale, since $\phi$ and $-\phi$ both give
  $\mathrm{MAC}{=}1$, so it measures mode-shape and direction alignment
  independently of amplitude~\cite{allemang1982mac}, and, being sign-invariant, it
  is complementary to sign agreement above. It is defined only for vector and modal
  fields such as displacement components and mode shapes. Scalar components such as
  stress or pressure leave it undefined, and it is reported as N/A for them.
\item[PRDC ($\uparrow$).] Precision, Recall, Density, and Coverage computed from
  the $k$-nearest-neighbor manifolds of the real and generated feature
  embeddings~\cite{naeem2020prdc}. Recall and Coverage collapsing while Precision
  holds is the signature of mode collapse.
\end{description}

For 3D geometry generation we additionally report standard point-cloud
distributional and fidelity measures. \emph{FPD} ($\downarrow$) is the Fr\'echet
distance between Gaussian fits to PointNet\texttt{++} feature embeddings of the
real and generated sets, the point-cloud analogue of FID. \emph{MV-FID}
($\downarrow$) is a multi-view FID that renders each cloud to sixteen fixed
depth-buffer views and computes the image FID over the rendered views, capturing
silhouette and surface fidelity that a single global feature distance can miss.
\emph{MS-SSIM} ($\downarrow$) is additionally computed pairwise among those
rendered views as a diversity proxy, where lower values indicate more varied
generated shapes, following the same repurposed direction convention used in 2D
generation below.
For set-level geometric fidelity we adopt the standard point-cloud generation
procedure under Chamfer distance~\cite{achlioptas2018pointcloud,yang2019pointflow}.
\emph{MMD-CD} ($\downarrow$), the minimum matching distance, averages over each
reference cloud the Chamfer distance to its nearest generated cloud, measuring
sample fidelity. \emph{COV-CD} ($\uparrow$), the coverage, is the fraction of
reference clouds that are the nearest neighbor of at least one generated cloud,
measuring mode coverage and thus directly penalizing collapse. \emph{1-NNA-CD}
($\downarrow$) is the leave-one-out $1$-nearest-neighbor two-sample
accuracy~\cite{yang2019pointflow}, which classifies each cloud in the pooled real
and generated set by the label of its nearest neighbor. Its ideal value is $0.5$,
at which real and generated are indistinguishable, and higher values indicate a
distinguishable generator that is under-fit or mode-collapsed. In the small-data
regime studied here every model scores above $0.5$, so we treat 1-NNA-CD as
lower-is-better. These set-level metrics replace an earlier pairwise Chamfer, EMD,
and F-Score summary that matched each generated cloud to an arbitrary reference and
hence rewarded a mode-collapsed generator emitting a single near-central shape.

For 2D depth regression we additionally report the standard monocular-depth
metrics AbsRel, sqRel, and threshold accuracy $\delta{<}1.25$~\cite{eigen2014depth},
computed on the foreground mask, as described in
Section~\ref{sec:results:pertask}.

For 2D generation, three otherwise-standard reconstruction metrics are repurposed
as unconditional-generation proxies, with their directions set accordingly. LPIPS
and MS-SSIM are computed pairwise among the generated samples as diversity
proxies, so higher LPIPS ($\uparrow$) and lower MS-SSIM ($\downarrow$), meaning
broader coverage and less mode collapse, are preferred, deliberately opposite to
their usual reconstruction-similarity convention. PSNR ($\uparrow$) is a coarse
whole-set fidelity proxy, higher when the generated pixel statistics sit closer to
the real set on these aligned, centered engineering renders. It is one of eight
jointly-ranked metrics rather than a headline score.

\section{BenchRank Ranking Robustness}
\label{app:robustness}

BenchRank's hyperparameters are fixed by construction, as described in
Section~\ref{sec:method:benchrank}. Here we show that the resulting rankings are
largely insensitive to them, using only post-processing of the committed per-run
results, with no retraining, on the same basis as the paper's leaderboards, namely
the published per-task seed basis and the same model pool, which adds the official
AB-UPT on DrivAerML to the published set as Section~\ref{sec:method:models}
describes. For seven representative task leaderboards, each at
all four data scales, we recompute the quality-view ranking under one-at-a-time
perturbations of the three knobs. These are redundancy strength
$\alpha\in\{0.5,2.0\}$ with default $1.0$, decisive-gap fraction
$\tau_d\in\{0.5,0.7\}$ with default $0.6$, and PageRank damping
$\delta\in\{0.80,0.90\}$ with default $0.85$. We then measure the Kendall
$\tau$-b rank correlation against the default ranking, reported in
Table~\ref{tab:robustness}. Across all $168$ task, scale, and perturbation
comparisons the mean $\tau$-b is $0.98$, with a worst case of $0.79$, and the
top-ranked model is unchanged in $160$ of $168$ cells, or $95\%$. The ranking is
most stable to damping at mean $\tau{=}1.00$ and to $\alpha$ at $0.99$, and
most sensitive to the decisive-gap fraction at mean $0.94$, which accounts for six
of the eight rank-1 flips. Five of those eight fall on one pair of boards, 2D field
prediction at M and L, where the default top two are separated by $0.01\%$ and
$0.21\%$ of total score, so the ranking there is a near-tie that any reweighting can
turn over. The other three sit on the two pooled multi-dataset scalar boards, where
the decisive-gap fraction changes how strongly per-dataset evidence is debiased
before the cross-dataset aggregation.

A fourth design choice is examined separately because it is not a hyperparameter,
namely the rule that rescales each cell's centrality onto a reported interval before the
cells are aggregated. Within a cell every candidate rule is an increasing affine map of
the same centrality vector, so no strict within-cell reversal occurs in any of the $260$
cells and such a rule can act only through the cross-dataset geometric mean. Replacing
the reported interval, whose lower end tracks the realized dominance-graph density, with
a fixed interval leaves the pooled ranking at mean $\tau$-b $0.98$ and changes the
top-ranked model in $1$ of $28$ task-scale boards, and replacing it with a scale-free
centrality ratio gives mean $0.95$ and changes the same $1$. That board is 2D field
prediction at L, where the default top two sit within $0.21\%$ of total score.

The training run is examined on the same footing. Recomputing every board with each task
placed on the mean of its repeated runs leaves the pooled ranking at mean $\tau$-b $0.97$
with a worst case of $0.73$ and changes one top-ranked model, in $1$ of $28$ task-scale
boards. That board is 3D generation at M, where the change does not reverse a decisive
result but dissolves one. The two leading models sit $42.33\%$ apart before averaging and
within $0.38\%$ of each other after, so the recomputed board reports them as effectively
tied rather than reversing a settled order.

The size of the run-to-run variation behind these figures shows up directly in the
per-cell spreads. Take 2D generation. Over its $108$ cells the standard deviation of FID
across the three runs is $3.9\%$ of the cell's own value at the median and $10.2\%$ at the
ninetieth percentile. On the reported boards the leading model is ahead of the runner-up by
$48.5\%$ of total score at the median, against a run-to-run spread in that score of
$1.2\%$ at the median and $6.1\%$ at the ninetieth percentile. Ranking each of the three
runs on its own returns the published top model on all twelve of that task's board-and-run
combinations.

As an external check, we also compare the default BenchRank ranking against two
simpler aggregators computed from the same cross-dataset metric matrix, a min--max
normalized mean and a Borda count. BenchRank is positively but not perfectly
correlated with both, at mean $\tau$-b $0.72$ and $0.75$, which confirms that its
debiased, dominance-graph aggregation is not a relabeling of a naive average. The
agreement is lowest exactly where naive aggregation is known to fail, namely the
pooled, unequal-coverage 3D-field board and the multi-dataset 1D-scalar tasks with
near-mean-collapse cells, which is precisely where a redundancy-debiased ranking is
designed to differ. This study is post-hoc. The seed basis is fixed per task rather
than per model, because ranking averaged models against un-averaged ones inside one
board would remove the run-to-run luck from some competitors and not others, as
Section~\ref{sec:conclusion} sets out. We did not test an Elo-style
aggregator. The analysis is
reproducible from the released per-run scores by re-running the ranking procedure
under the stated perturbations, except for the AB-UPT rows of the 3D-field boards,
which its research license withholds from the release.

\begin{table}[h]
  \centering\footnotesize
  \caption[BenchRank ranking robustness under hyperparameter perturbation]%
  {BenchRank ranking robustness, with all four scales pooled per task.
  Columns give the Kendall $\tau$-b of the quality ranking against the default
  under hyperparameter perturbation as min and mean, the share of the four scales
  crossed with the six perturbations whose top model is unchanged, labeled
  ``Rank-1 stable'', and $\tau$-b against a normalized-mean and a Borda
  aggregator.}
  \label{tab:robustness}
  \resizebox{\coltabw}{!}{\begin{tabular}{@{}l c c c c c@{}}
\toprule
Task & \#(scale) & Kendall $\tau$ & Rank-1 & $\tau$ vs & $\tau$ vs \\
 & cells & (min / mean) & stable & norm-mean & Borda \\
\midrule
2D generation & 4 & 0.94\,/\,1.00 & 24/24 & 0.76 & 0.82 \\
3D generation & 4 & 0.87\,/\,0.98 & 24/24 & 0.90 & 0.90 \\
2D field prediction & 4 & 0.79\,/\,0.95 & 19/24 & 0.73 & 0.77 \\
2D scalar prediction & 4 & 0.90\,/\,0.99 & 23/24 & 0.81 & 0.86 \\
3D scalar prediction & 4 & 0.96\,/\,0.99 & 24/24 & 0.94 & 0.96 \\
3D field prediction & 4 & 0.85\,/\,0.97 & 24/24 & 0.57 & 0.66 \\
1D scalar prediction & 4 & 0.83\,/\,0.96 & 22/24 & 0.29 & 0.31 \\
\bottomrule
\end{tabular}
}
\end{table}

\ifdefined\arxivbuild
  \bibliographystyle{unsrtnat}
\else
  \bibliographystyle{asmejour}
\fi
\bibliography{refs}

\end{document}